%% file: main.tex
\pdfoutput=1

\documentclass[11pt]{article}

\usepackage{ACL2023}

\usepackage{times}
\usepackage{latexsym}

\usepackage[T1]{fontenc}

\usepackage[utf8]{inputenc}

\usepackage{microtype}

\usepackage{inconsolata}

\usepackage{color, colortbl}
\definecolor{azure}{rgb}{0.0, 0.5, 1.0}

\input{packages.tex}

\title{\textsc{MYTE}: \underline{M}orphology-Driven B\underline{yte} Encoding for Better and Fairer Multilingual Language Modeling}

\author{Tomasz Limisiewicz$^{1\dagger*}$ \quad  Terra Blevins$^2$ \quad Hila Gonen$^2$ \\ \bf Orevaoghene Ahia$^2$ \quad Luke Zettlemoyer$^2$ \\
$^1$Faculty of Mathematics and Physics, Charles University in Prague \\
$^2$Paul G. Allen School of Computer Science and Engineering, University of Washington }

\begin{document}
\maketitle
\begingroup\def\thefootnote{$\dagger$}\footnotetext{Correspondence to {\tt limisiewicz@ufal.mff.cuni.cz}}\endgroup
\begingroup\def\thefootnote{*}\footnotetext{Work done while visiting the University of Washington}\endgroup

\input{sections/00_abstract}

\input{sections/01_introduction}

\input{sections/02_utf8_background}
\input{sections/03_methodology}

\input{sections/04_morphological_bytes}

\input{sections/05_myt5}

\input{sections/06_related_work}

\input{sections/07_conclusion}
\input{sections/08_es_limitations}

\section*{Acknowledgements}
We would like to thank Jana Straková and Zdeněk Žabokrtský for helpful feedback on this project. 
We are thankful to Google for providing free computation quotas through the TPU Research Cloud program.
Tomasz Limisiewicz acknowledges the support of grant 338521 of the Charles University Grant Agency, a Fellowship from Paul G. Allen School, and the Mobility Fund of Charles University.

\bibliography{custom_new, anthology_new}
\bibliographystyle{acl_natbib}
\clearpage
\appendix

\input{sections/91_morphological_analysis}
\input{sections/92_supplementary_results}

\input{sections/93_technical}

\end{document}

%% file: packages.tex
\usepackage{subcaption}
\usepackage{graphicx}
\usepackage{booktabs}
\usepackage{multirow}
\usepackage{amsmath}
\usepackage{bbold}
\usepackage{afterpage}
\usepackage{makecell}

\usepackage{tikz}
\newcommand\bytebox[2][]{\tikz[overlay]\node[rectangle, , inner sep=1pt, anchor=text, rounded corners=1mm, draw = blue!80, ultra thick, minimum size=12pt,#1] {\textcolor{blue!80}{#2}};\phantom{#2}}

\newcommand\mytebox[2][]{\tikz[overlay]\node[rectangle, , inner sep=1pt, anchor=text, rounded corners=1mm, draw = bottlegreen!80, ultra thick, minimum size=12pt,#1] {\textcolor{bottlegreen!80}{#2}};\phantom{#2}}

\newcommand{\utf}[0]{\emph{UTF-8}}
\newcommand{\ours}[0]{\textsc{MYTE}}
\newcommand{\byt}[0]{ByT5}
\newcommand{\myt}[0]{MyT5}
\newcommand{\xu}[0]{\textsc{Xtreme-Up}}

\definecolor{bottlegreen}{rgb}{0.0, 0.416, 0.306}
\definecolor{table_red}{HTML}{FF5555}
\definecolor{table_green}{HTML}{55BB55}
\definecolor{table_blue}{HTML}{5555FF}
\definecolor{table_gray}{HTML}{555555}

%% file: sections/00_abstract.tex
\begin{abstract}
A major consideration in multilingual language modeling is how to best represent languages with diverse vocabularies and scripts.
Although contemporary text encoding methods cover most of the world's writing systems, they exhibit bias towards the high-resource languages of the Global West.
As a result, texts of underrepresented languages tend to be segmented into long sequences of linguistically meaningless units.
To address the disparities, we introduce a new paradigm that encodes the same information with segments of consistent size across diverse languages.
Our encoding convention (\ours{})
is based on morphemes, as their inventories are more balanced across languages than characters, which are used in previous methods.
We show that \ours{} produces shorter encodings for \textit{all 99} analyzed languages, with the most notable improvements for non-European languages and non-Latin scripts.
This, in turn, improves multilingual LM performance and diminishes the perplexity gap throughout diverse languages.

\end{abstract}

%% file: sections/01_introduction.tex
\section{Introduction}


Multilingual language models have become the state-of-the-art solution for performing tasks on a wide range of languages \cite{devlin-etal-2019-bert, conneau-etal-2020-unsupervised, xue-etal-2021-mt5}.
However, it is challenging to ensure high performance for all languages due to differences in data availability, especially for the long tail of low-resource languages \cite{malkin-etal-2022-balanced}. This challenge is compounded by choices of how words are represented during tokenization; past studies have shown that multilingual models either cannot accurately represent texts in rare languages \cite{pfeiffer-etal-2021-unks} or do so via over-segmentation, which is detrimental both to model performance and inference cost \cite{petrov_language_2023, ahia_all_2023}. 

\input{figures/myte_example}

Byte-level models aim to solve these challenges. Rather than words or subword tokens, they use byte-level text representations that achieve high coverage \cite{xue_byt5_2022}, as common encodings such as UTF-8 support most of the world's scripts.
Nevertheless, the over-segmentation problem still exists even at the byte level, as byte sequences for single characters are overly long for many non-Latin script languages \cite{arnett2024bit}. 
This problem has an immense effect on modeling these scripts in NLP systems, as operating on longer sequences significantly increases the computation costs of training and inference in models, while also making learning less sample efficient. 
Furthermore, the billing for APIs such as ChatGPT (\url{openai.com/chatgpt})
is often associated with the segmented sequence length, disadvantaging speakers of specific languages \cite{ahia_all_2023}. 
 

In this work, we propose a novel method to derive byte representations of text, enabling equitable segmentations across languages and scripts.
In our approach, we replace the current convention of assigning byte codes to characters with a morphology-driven approach, as morphemes\footnote{In this work, the usage of term ``morphemes'' encompasses both ``morphemes'' and ``morphs''. Some linguistic theories use the term ``morph'' for specific textual realizations of abstract ``morphemes''. For instance, in English, \textit{es} as in \textit{fox\underline{es}} and \textit{s} as in \textit{cat\underline{s}} are two distinct ``morphs'' of a plurality ``morpheme''. For an in-depth discussion about these two terms, see Section 4 of \citet{zabokrtsky-etal-2022-towards}} are more informatively comparable constituents of text across languages than characters \cite{cotterell_are_2018}.
Specifically, we introduce a novel algorithm for representing text as byte sequences that is based on unsupervised morphological segmentation \cite{smit-etal-2014-morfessor}.
We demonstrate that our new paradigm for byte representation improves the segmentation of diverse languages of various scripts and morphological inventories. Furthermore, the segmentation of parallel sentences across languages converges to comparable lengths. 


We test our method's effectiveness in creating equitable text representation -- representations that given parallel texts have similar encoded sequence lengths.
We then evaluate the applicability of the method to multilingual language modeling across 99 typologically diverse languages.

Our contributions can be summarized as follows: (a) We propose a novel byte-encoding method that is morphologically driven; (b) We show empirically that the resulting representations are more equitable across languages than vanilla byte, character, or subword segmentation; (c) We analyze the typical lengths of these representations and show decreased 
sequence length across all analyzed languages, significantly reducing computation cost and benefiting non-Latin script languages the most; (d) We train a language model with our new representation scheme and demonstrate that it maintains balanced and better LM performance across diverse languages and exhibits faster inference speed. This improvements holds across different model scales. 
Our models match SOTA \byt{} performance across multiple tasks for diverse low-resource languages while being more efficient in training and inference.



We will release our code and models to facilitate further research in this direction.


%% file: figures/myte_example.tex
\begin{figure}[!tb]
    \includegraphics[width=\linewidth]{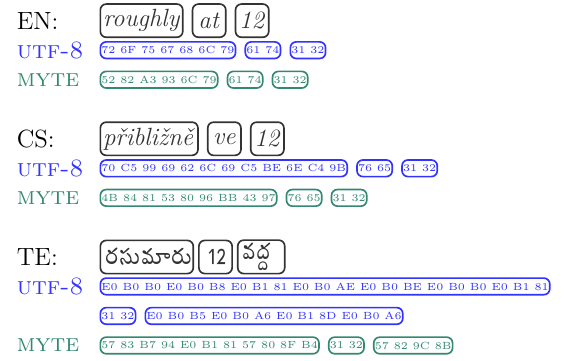}
    \caption{The same phrase is spelled in three languages: English, Czech, and Telugu. \textcolor{blue!80}{\utf{}} byte encoding of the phrase is shown in blue, while \textcolor{bottlegreen!80}{\ours{}} in green underneath. \ours{} achieves higher encoding compression, especially for texts using diacritics or non-Latin script.}
    \label{fig:myte_example}
\end{figure}

%% file: sections/02_utf8_background.tex
\section{Background: UTF-8 Bytes}

\label{sec:utf8-bg}

The vast majority of texts online\footnote{\url{https://w3techs.com/technologies/overview/character_encoding}} are represented as bytes via \utf{} convention, which is defined by the Unicode Standard \cite{the_unicode_consortium_unicode_2011}.
In \utf{}, each character (or codepoint) is represented as a sequence of one to four bytes.
Due to the gradual development of communication standards, \utf{} first allocated one-byte representation ASCII symbols, which cover primary Latin-script characters (see \bytebox[]{00} to \bytebox[]{7F} in Figure \ref{fig:utf8-codespacr}).
Other characters are represented as multi-byte codes starting with a byte from range \bytebox[]{C2} to \bytebox[]{F4} denoting the number of bytes in the codepoint and followed by continuation bytes from range \bytebox[]{80} to \bytebox[]{BF}.

In \utf{} convention, characters in non-Latin alphabetic scripts (Cyrillic, Armenian, Georgian), diacritics, and abjads\footnote{Abjads are writing scripts that do not denote vowels, e.g., Hebrew, Arabic.} 
usually have two-byte codes, while the byte length increases to three or four for Brahmic abugidas\footnote{Abugidas are scripts representing consonant-vowel as one character, typical to the Indian Subcontinent and South-East Asia, e.g., Devanagari, Bengali.} and CJK (Chinese, Japanese, Korean) logographs.
As a result, the granularity of byte codes varies significantly across languages; this means that texts conveying the same information across languages tend to be represented by byte sequences of significantly different lengths \cite{arnett2024bit}.


%% file: sections/03_methodology.tex
\input{figures/utf8_codespace}

\section{Method: Morphology-Driven Bytes}


As discussed in the prior section and shown in Figure~\ref{fig:myte_example}, \utf{} convention produces longer byte sequences for some languages due to the development choices.
To make byte representation more equitable, we introduce an encoding paradigm that aims to assign byte codes of similar lengths to morphemes across languages.
We base our encoding scheme on morphological analysis because morphemes are 
the shortest meaningful constituents and are independent of the writing convention \cite{haspelmath_understanding_2010}.
We assume that the number of morphemes in sentences with the same information load is more balanced across languages than the number of characters, bytes, or tokens.
Thus, we enforce balanced segmentation granularity across languages.

An alternative approach to encoding morphological representations would be treating the union of multilingual morpheme inventories across languages as one large subword vocabulary.
To cover the morphemes of many languages in this manner, the vocabulary would be much larger than the ones usually applied to models.\footnote{The proposed \ours{} encoding offers capacity for 2,130,432 of variable length codepoints. It is considerably more than in any of the commonly used subword vocabularies. For reference, large vocabulary XLM-V model allocates 1 million subwords \cite{liang-etal-2023-xlm}.} This would incur additional computational costs and, similar to other subword representations, would likely not generalize well to new, unseen languages.

\subsection{Morphological Analysis}
\label{sec:morph-analysis}

We train an unsupervised morphological analyzer, Morfessor \cite{smit-etal-2014-morfessor} on lexicons derived from whole Wikipedia articles in 99 languages. 
The morphological analysis is performed on each of the languages separately to balance the number of morphemes per language, regardless of data resourcefulness.
For each language, we derived a set of 4096 morphemes; the number was chosen to balance segmentation granularity across languages.
For each morpheme, we save its score, defined as the hypothetical loss reduction of the Morfessor model if the morpheme had not been included in the set.
We take the union of sets across languages to obtain a multilingual morpheme inventory.
The details of lexicon preparation and the usage of Morfessor are in Appendix~\ref{sec:app-morphology}.


\input{tables/script_groups}

\subsection{Enriching Byte Representation with Morphology} 
\label{sec:morphology-enriching}

To alleviate \utf{} inefficiencies, we propose a systematic rearrangement of byte codepage.
We free 26 bytes ( \bytebox[]{41} to \bytebox[]{5A} ) by decomposing capital letter codes into lowercase letters and capitalization markers. The first byte from this range (~\bytebox[]{41} ) is repurposed as a capitalization marker.
The remaining 25 bytes are freed space used to store morphemes. 

Our method takes the sequences of \utf{} bytes and transcodes them into shorter sequences using the vocabulary of the same size, i.e. 256, as depicted in Figure~\ref{fig:myte_example}. We apply the following steps to transcode \utf{} sequences to \ours{} encodings:
\begin{enumerate}
    \item We use \utf{} as base encoding of text.
    Then, the byte sequences are transcoded from left to right, merging morpheme sequences and replacing them as dedicated codepoints described in the following points.
    \item The morphemes are grouped by scripts as shown in Table~\ref{tab:script_groups}.
    Codepoints of multiple scripts within a single morpheme are assigned to the second cluster (Mixed script).
    \item The morphemes are ranked based on their Morfessor score defined in Section~\ref{sec:morph-analysis}.
    \item We assign multibyte codepoint for each of the morphemes analogously to the \utf{} convention (see Section~\ref{sec:utf8-bg}).
    Specifically, the first byte denoting the beginning of the morphological codepoint is assigned from the freed range (~\mytebox[]{42} - \mytebox[]{5A}~) based on the morph's inclusion in one of the script groups.
    It is followed by continuation bytes from the $64$ element range \mytebox[]{80} ~-~\mytebox[]{BF}, as in \utf{} convention.
    The $64$ morphemes with the highest score are saved as two-byte codepoints, following $64^2=4096$ as three-byte codepoints; the remaining morphemes are saved as up to $64^3=262,144$ four-byte codepoints. The capacity for new codepoints was not exhausted for any script group.
\end{enumerate}


%% file: figures/utf8_codespace.tex
\begin{figure}[!tb]
    \centering
    \includegraphics[width=\linewidth]{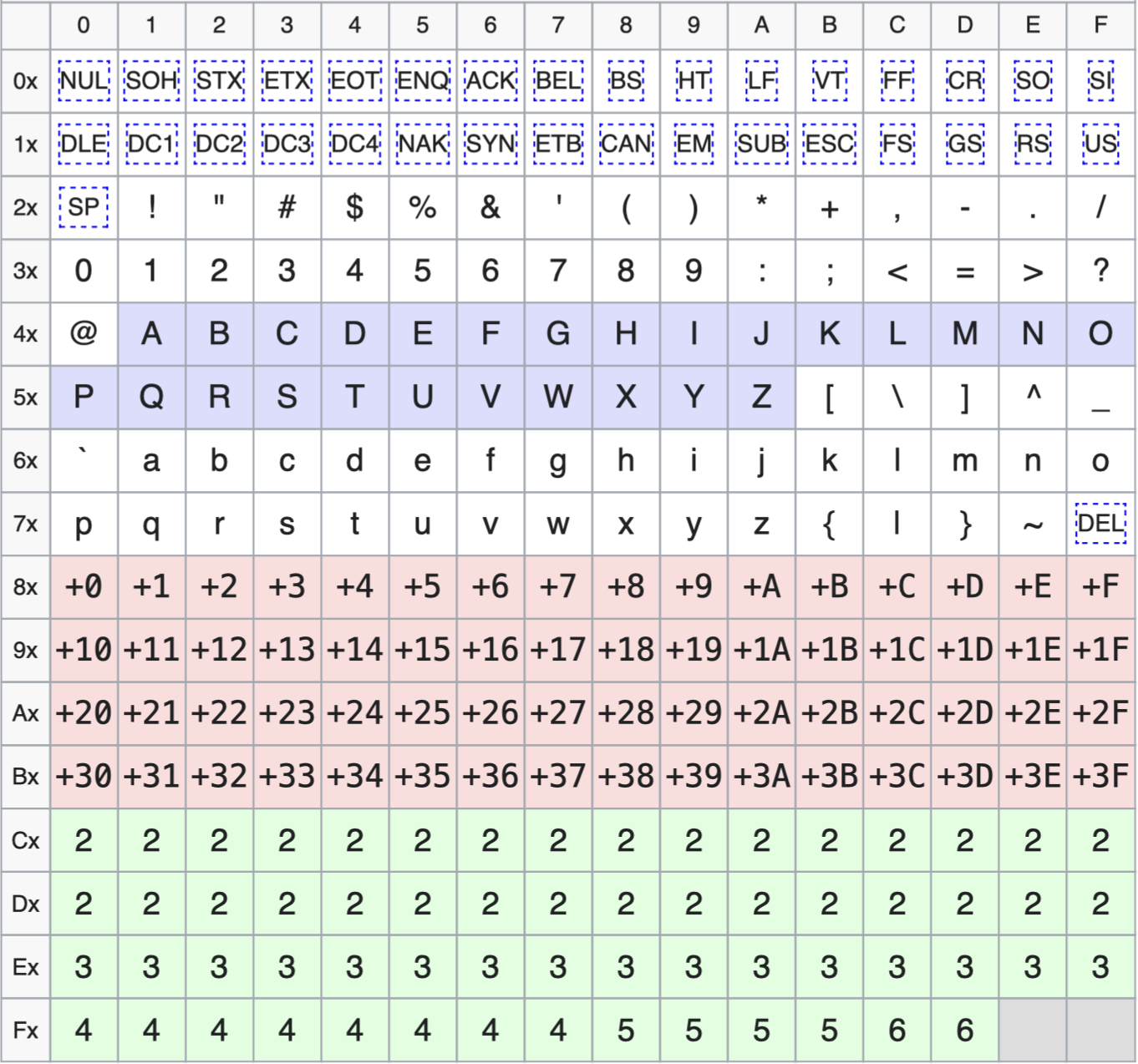}
    \caption{UTF-8 codepage (inspired by the visualizations from: \url{en.wikipedia.org/wiki/UTF-8}).
    Each row contains 16 bytes with the same leading hexadecimal digit. 
    Bytes in the range \bytebox[]{C2} - \bytebox[]{F4} are \textcolor{table_green}{leading bytes}. They mark the beginning of a multibyte code of the length shown in each cell.
    Bytes in the range \bytebox[]{80} - \bytebox[]{BF} are \textcolor{table_red}{continuation bytes}, which follow a leading byte in multibyte codes. Bytes \bytebox[]{FE} and \bytebox[]{FF} are \textcolor{table_gray}{unused}. Range \bytebox[]{41} - \bytebox[]{5A} encodes \textcolor{table_blue}{Latin capital letters}. In \ours{}, these characters are decomposed to free space used to encode morphemes. }
    \label{fig:utf8-codespacr}
\end{figure}

%% file: tables/script_groups.tex
\begin{table}[!tb]
\small
\begin{tabular}{@{}lp{0.55in}p{1in}lll@{}}
\toprule

\multirow{2}{*}{ID} & \multirow{2}{*}{Group} & \multirow{2}{*}{Unicode Script(s)}                                             & \multicolumn{3}{c}{Leading Byte}   \\ \cmidrule(l){4-6} 
                          &                             &                                                                              & 2 b            & 3 b            & 4 b            \\ \midrule
0                         & Latin                       & Latin                                                                        & \mytebox[]{42} & \mytebox[]{4A} & \mytebox[]{52} \\ \\
1                         & Common                      & Mixed, Common, Inherited, Unkown                                             & \mytebox[]{43} & \mytebox[]{4B} & \mytebox[]{53} \\ \\
2                         & Non-Latin Alphabetic        & Greek, Cyrillic, Armenian, Georgian                                          & \mytebox[]{44} & \mytebox[]{4C} & \mytebox[]{54} \\ \\
3                         & Abjads                      & Hebrew,  Arabic,  Syriac,  Thaana,  Tifinagh                                 & \mytebox[]{45} & \mytebox[]{4D} & \mytebox[]{55} \\ \\
4                         & Abugidas North              & Devanagari, Gurmukhi, Gujarati, Oriya, Bengali, Sinhala, Tibetan             & \mytebox[]{46} & \mytebox[]{4E} & \mytebox[]{56} \\ \\
5                         & Abugidas South              & Telugu, Kannada, Tamil, Malayalam, Thai,  Lao, Myanmar,  Tai, Tagalog, Khmer & \mytebox[]{47} & \mytebox[]{4F} & \mytebox[]{57} \\ \\
6                         & CJK  & Hangul, Han, Yi, Katakana, Hiragana, Bopomofo                                & \mytebox[]{48} & \mytebox[]{50} & \mytebox[]{58} \\ \\
7                         & Other                       & Remaining scripts                                                            & \mytebox[]{49} & \mytebox[]{51} & \mytebox[]{59} \\ \bottomrule
\end{tabular}
\caption{Groups of scripts with the initial bytes for their morphological blocks. The groups were selected to balance the number of covered languages with similar writing systems.}
\label{tab:script_groups}
\end{table}

%% file: sections/04_morphological_bytes.tex
\section{Equitable Multilingual Segmentation with \ours{}}


We first analyze the properties of our proposed morphology-driven encoding. 
Following the setting of \newcite{petrov_language_2023}, we measure whether \ours{} produces the segmented sequences of comparable length across languages. 

We compute parity across languages using the multi-parallel corpus Flores 200 \cite{nllb2022}.
Parity is defined as $|t(s_l)|/|t(s_{en})|$, where $s_l$ and $s_{en}$ stand for parallel sentences in language $l$ and in English, respectively. $|t(s)|$ is the length of sequence $s$ with segmentation method $t$.

We compare the \ours{} encoding from Section~\ref{sec:morphology-enriching} to several baselines of common input representation: (a) Vanilla byte-level encoding via UTF-8; (b) Character-level encoding; (c) Subwords produced by Sentencepiece algorithm \cite{kudo-richardson-2018-sentencepiece}. In comparison, we focus on the equitability of sequence lengths produced by the methods for diverse languages.

Furthermore, we compare our morphological byte encoding sequence compression rate against the \utf{} convention. Compression is essential for an effective text representation as it affects NLP systems' efficiency and usage cost \cite{ahia_all_2023}.
Finally, we check whether our method more effectively compresses languages and scripts unseen in \ours{} algorithm described in Section~\ref{sec:morphology-enriching}.

\subsection{Results}
\label{sec:results-sequences}

\paragraph{\ours{} is Equitable across Languages}
The comparison of sequence length across parallel sentences in Flores 200 is shown in Figure~\ref{fig:bseq-comparison}. Our representation is more balanced across languages than the original \utf{} bytes. There are still four languages with observably higher code lengths (e.g., Greek, Vietnamese, Punjabi, Khmer).
However, \ours{} encoding still improves their parity to English such that it is much lower than outlier languages in \utf{} (1.7 vs. 3.5 in the worst-case languages, respectively).

Figure~\ref{fig:parity} shows that \ours{} representations are more balanced in parity scores across languages than subword tokenization. In particular, we improve on the long tail of languages over-segmented either in byte or subword encoding. 
The parties closest to \ours{} are obtained by character representation. 
However, the set of all Unicode characters is larger by orders of magnitude than the number of unique bytes used in \ours{} (149,878 vs. 254).

\input{figures/parity_comparison}

\input{figures/byte_sequences_comparison}

\input{tables/aggregated_parities}

\paragraph{\ours{} Encoding Compresses Text Representation}
The encoded sequence lengths are decreased with \ours{} encoding for all languages, as depicted in Figure~\ref{fig:bseq-compressions}.
The rate of compression varies from ~1\% for Vietnamese and Chinese to almost 70\% for Burmese. As seen in Table~\ref{tab:aggregated_parities}, the highest compression is obtained for low-resource languages with non-Latin scripts.
Notably, this group of languages is the most susceptible to over-segmentation in \utf{} encoding.

\paragraph{Findings Generalize to Unseen Languages but not Unseen Scripts}
In Table~\ref{tab:aggregated_parities}, we observe that a decrease in sequence length and parity applies to five low-resource languages not considered in constructing \ours{} representation, referred to as \emph{unseen languages}.
One exemption from the rule is Santhali, written in \emph{unseen} Ol Chiki script, for which we do not observe a change in the encoded sequence length.
This observation highlights the importance of considering a wide range of languages and scripts when constructing morpheme inventories.
Importantly, \ours{} did not reach a capacity of available byte codepoints, and thus, the method can be extended to additional languages. The complete results for \emph{unseen} languages and scripts are shown in Appendix~\ref{sec:app-results}.

%% file: figures/parity_comparison.tex
\begin{figure}[!tb]
    \centering
    \includegraphics[width=\linewidth]{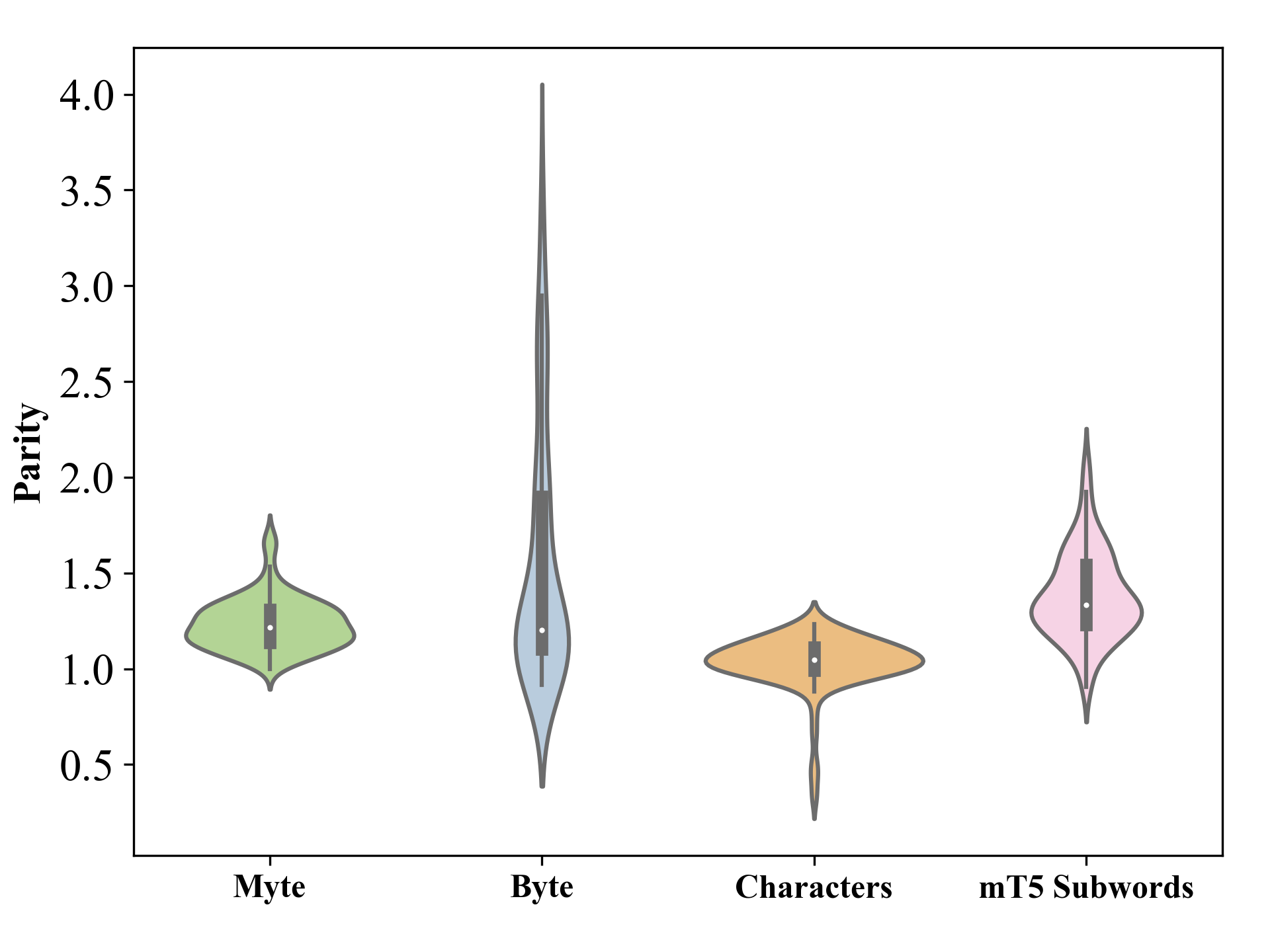}
    \caption{Boxplot aggregating parity against English for three segmentation methods: \ours{}, \utf{}, characters, and subword tokens from mT5 tokenizer \cite{xue-etal-2021-mt5}. Parities were computed on multi-parallel Flores 200 corpus. }
    \label{fig:parity}
\end{figure}

%% file: figures/byte_sequences_comparison.tex
\begin{figure}[!tb]
    \centering
    \begin{subfigure}[b]{\linewidth}
        \includegraphics[width=\textwidth]{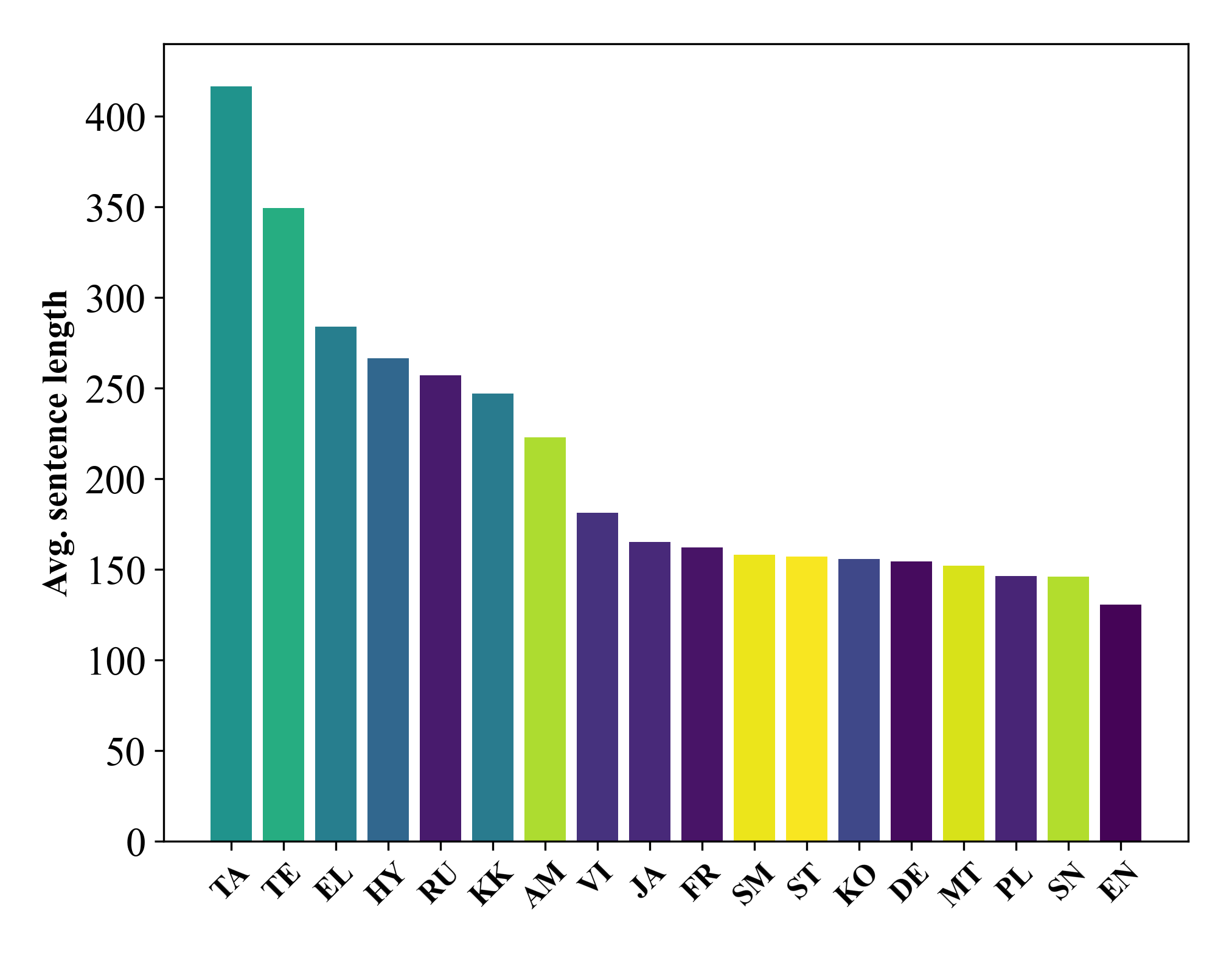}
        \caption{\utf{}}
    \end{subfigure}
    \begin{subfigure}[b]{\linewidth}
        \includegraphics[width=\textwidth]{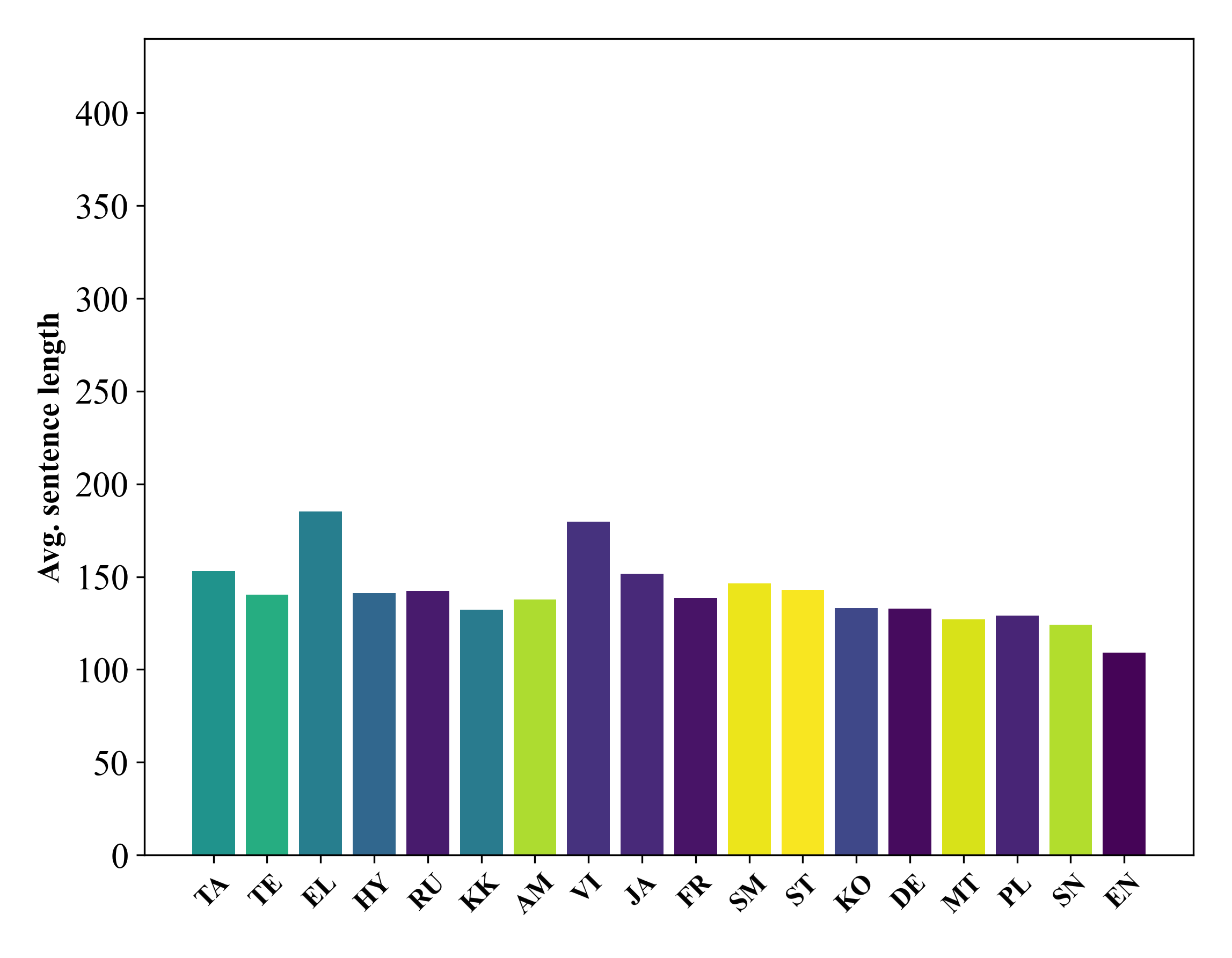}
        \caption{\ours{}}
    \end{subfigure}
    \begin{subfigure}[b]{\linewidth}
        \includegraphics[width=\textwidth]{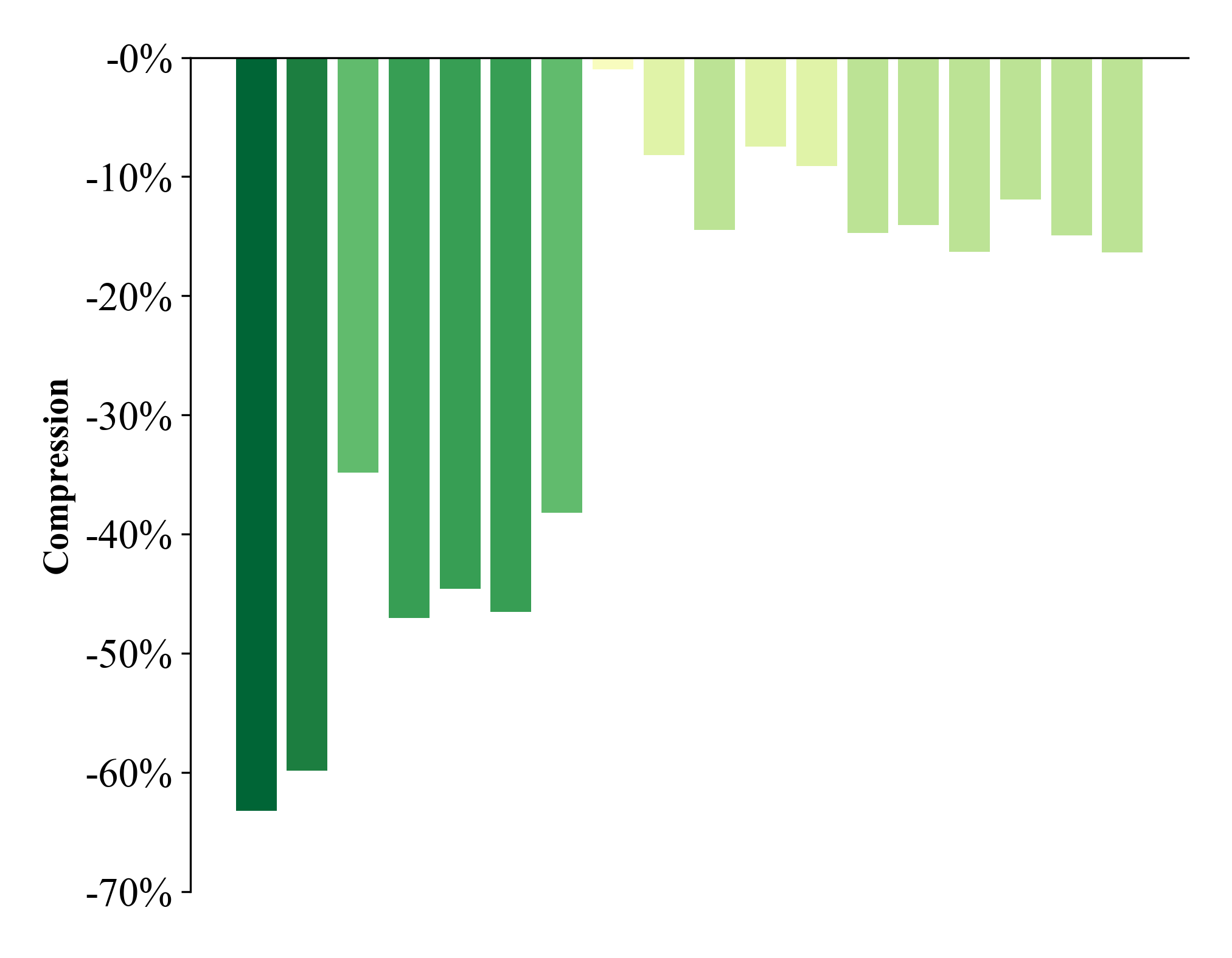}
        \caption{Sequence compression}
        \label{fig:bseq-compressions}
    \end{subfigure}
    \caption{Average byte sequence lengths of parallel sentences from Flores 200 encoded by a) \utf{} and b) \ours{}. Figure c) depicts the percentage by which the latter sequences are shorter than the former. Results for all the languages can be found in Appendix~\ref{sec:app-results}.
    }
    \label{fig:bseq-comparison}
\end{figure}

%% file: tables/aggregated_parities.tex
\begin{table}[th]
\centering
\small
\begin{tabular}{lccccc}
\toprule
{} & \multicolumn{2}{c}{Byte} & \multicolumn{2}{c}{Myte} & Comp. \\ \cmidrule(l){2-3} \cmidrule(l){4-5}  
{} & Parity & Len. & Parity & Len. & \\
\midrule
English       &   1.00 & 131 &   1.00 & 109 &        16\% \\ \midrule
Latin HR      &   1.14 & 149 &   1.18 & 129 &        14\% \\
Latin LR      &   1.12 & 147 &   1.18 & 128 &        12\% \\
$\neg$Latin HR  &   1.62 & 212 &   1.29 & 141 &        29\% \\
$\neg$Latin LR  &   2.33 & 305 &   1.33 & 145 &        50\% \\ \midrule 
\textbf{Seen} &   1.56 & 204 &   1.24 & 135 &        26\% \\ \midrule \midrule
Unseen Lang   &   1.50 & 196 &   1.27 & 138 &        23\% \\
Unseen Script &   2.80 & 365 &   3.35 & 365 &        0\% \\  \midrule
\textbf{Unseen} &   1.72 & 224 &   1.61 & 176 &        19\% \\
\bottomrule
\end{tabular}
\caption{Averaged sequence length and corresponding parities to English of \utf{} and \ours{}. We aggregated results for languages used in morphological adaptation (i.e., \emph{Seen}) by their script (Latin vs. Non-Latin) and resourcefulness (HR: high resource, LR: low resource) based on categorization from \citet{joshi-etal-2020-state}. The last three rows present results for languages \emph{unseen} in morphological adaptation; all of them are low-resource. Shortened column headers: Len. -- Length, Comp. -- Compression.}
\label{tab:aggregated_parities}
\end{table}

%% file: sections/05_myt5.tex
\section{\myt{}: Language Modeling with \ours{}}

This section investigates the benefits of \ours{} as an encoding scheme for byte-level language modeling.
For that purpose, we have trained T5 language models on \ours{} representation.
We refer to these models as \textbf{My}te \textbf{T5} models, or \myt{} for short.

\input{figures/bpeb_time_delta}

\subsection{Training Details}

We base the architecture and implementation of our \myt{} model on the byte-level T5 model, i.e., \byt{} \cite{xue_byt5_2022}.
\byt{}, like other T5 models \cite{raffel_exploring_2019}, is an encoder-decoder Transformer model trained on predicting masked spans of texts. 
\byt{} operates on bytes instead of the subword tokenization in the standard T5 model, making it a suitable base model for our setting.


We pre-train three new instances of \ours{}-level models of different sizes: small (300M), base (582M), and large (1.23B parameters). For pre-training, we used the standard task of restoring corrupted spans from mC4 corpus \cite{raffel_exploring_2019}.
All the byte sequences are transcoded into morphologically-driven bytes.
We use Jax implementation, i.e., t5x repository \cite{roberts_scaling_2022}, and the same hyperparameters as in \byt{} \cite{xue_byt5_2022}.
The only difference from their training approach is that we pre-train for 250,000 steps rather than one million steps since we observe overfitting when training for more steps, especially on low-resource languages. \citet{chung2023unimax} similarly observed overfitting in multilingual T5 models caused by extensive duplications in the mC4 corpus, leading them to also train models for only 250,000 steps.
In evaluations, we compare against a reimplemented \byt{} instance trained for the same number of steps. 

\subsection{Experiments}
We compare the performance of the \myt{} and \byt{} models, focusing on three 
aspects: language modeling performance, efficiency, and downstream evaluation.

First, the multilingual language modeling performance of \myt{} -- how is it, and is it comparable across languages? 
Inspired by \newcite{cotterell_are_2018}, we use the Bit-per-English-Byte metric on the multi-parallel FLORES 200 corpus to control for the informativeness of evaluation sequences:

\begin{equation}
    BPEB = \frac{1}{|\mathbf{c}_{English, UTF}|+1}\sum_{i=1}^{|\mathbf{c}|+1}\log p(c_i|\mathbf{c}_{<i})
\end{equation}

$\mathbf{c}$ is a sequence of bytes (original \utf{} or \ours{}) with $c_i$ being the $i$-th byte.
For normalization, we use the number of \utf{} bytes in English sentence $\mathbf{c}_{English, UTF}$ for fair comparison across languages and representation methods. 
It is the main difference from perplexity, which is normalized by the sequence length and thus confounded by segmentation rates characteristic of individual languages and encodings.

Second, we compare inference times of text generation of \myt{} and \byt{}.
We expect a decrease in sequence length, as shown in the last section, will render up to a quadratic reduction of forward-pass time due to the quadratic complexity of attention computation.
For both aspects, we report the results on three scales of the model (small, base, and large). Unless stated otherwise, we present the results of the large model. 

Lastly, we compare models' performance on four tasks from the \xu{} benchmark \cite{ruder_xtreme-up_2023}: question answering, named entity recognition, semantic parsing, and translation from English.
In each task, we fine-tune the large models on the multilingual data of all languages for each task.
Fine-tuned models are evaluated on test data for low-resource languages, following \citet{ruder_xtreme-up_2023}.
The only exception is machine translation, where we fine-tune and evaluate on a subset of languages to reduce the computation cost.
The details of training and evaluation are provided in Appendix~\ref{sec:appendix-technical}.



\subsection{Results}
\paragraph{\myt{} Outperforms \byt{} in Language Modeling}
In Figure~\ref{fig:bpeb-delta}, our model outperforms \byt{}, producing lower (better) average $BPEB$ scores for all analyzed languages. 
The improvement is strongly negatively correlated with the compression rate discussed in the previous section. 
The gains are largest for languages using Abugidas (scripts representing consonant-vowel as one character, typical to the Indian Subcontinent and SE Asia) that tend to be shortened the most by \ours{} encoding.
On the other end of compression distribution, we still observe (smaller) improvement for Latin and CJK scripts.
This observation suggests that the \ours{} encoding's leverage is not constrained to shortening sequences, but it also uses codepoints that are easier to predict by a language model. 
\ours{} uses codepoints based on morphemes that are inherently meaningful language units in contrast to orthographic symbols, which are the backbone of the \utf{} convention.

\input{figures/lm_flores}
\input{figures/lm_flores_unseen}

\paragraph{Encoding in \myt{} Diminishes LM Performance Gap Across Languages}

Previous works have argued that some languages are more challenging to model due to their morphological properties \cite{cotterell_are_2018}.
In contrast, others suggest that LM performance is linked with how texts in specific languages are represented \cite{park_morphology_2021}. 
Our results in Figure~\ref{fig:flores-lm} support the latter view, as the predictability of the languages is balanced by using equitable underlying representation, i.e., \ours{} encoding. 
Specifically, we show that \myt{} achieves more balanced $BPEB$ across languages than \byt{}. 
As discussed in the previous section, the benefit is the starkest for languages prone to over-segmentation under \utf{}.
The smallest improvements of \myt{} are obtained for languages benefited by \ours{} to a lesser extent, as observed in Section~\ref{sec:results-sequences}: Greek and Vietnamese. 

In Figure~\ref{fig:flores-lm-unseen}, we observe that \myt{} outperforms \byt{} for languages unseen in morphological analysis, except for Sanatli, which also uses a distinct script. 

\input{tables/aggregated_modeling}

\paragraph{\myt{} is More Efficient at Scale than \byt{}}
As shown in Figure~\ref{fig:time-delta}, \myt{}'s inference time is shorter than that of \byt{} for almost all languages.
This behavior is mostly observed for Non-Latin script languages and can thus be attributed to the higher rates of compression observed when using the \ours{} encoding scheme (Figure \ref{fig:bseq-comparison}).
Furthermore, Table~\ref{tab:aggregated_modeling} demonstrates that \myt{}'s inference speed gains over \byt{} improve with model size, hinting that \ours{} will bring further efficiency gains when applied to models of larger scales.


\input{figures/xtreme_up_comparison}
\input{tables/xtreme_up_all}

\paragraph{\myt{} Performs End Tasks Faster than \byt{}}

As shown in Table~\ref{tab:xtreme_up}, \myt{} and \byt{} perform comparably (and better than baselines) on MT and NER.
While \myt{} outperforms \byt{} by 2 points on QA, the opposite is true for semantic parsing.
We hypothesize that in this case, the morphological prior encoded in \ours{} may confound semantic parsing fine-tuning, which requires a structured output starkly dissimilar to natural language.

For all the tasks, the inference of \myt{} is faster than \byt{} (Figure~\ref{fig:xtreme_up_comparison}), mirroring our observations on language modeling efficiency. 
However, we do not observe a consistent relationship between the change in end task performance and efficiency, contrasting with the earlier observed correlation between $\Delta$ of inference time and $BPEP$ in multilingual language modeling.

%% file: figures/bpeb_time_delta.tex
\begin{figure*}[!tb]
\centering
    \begin{subfigure}[b]{0.44\linewidth}
        \includegraphics[width=\textwidth]{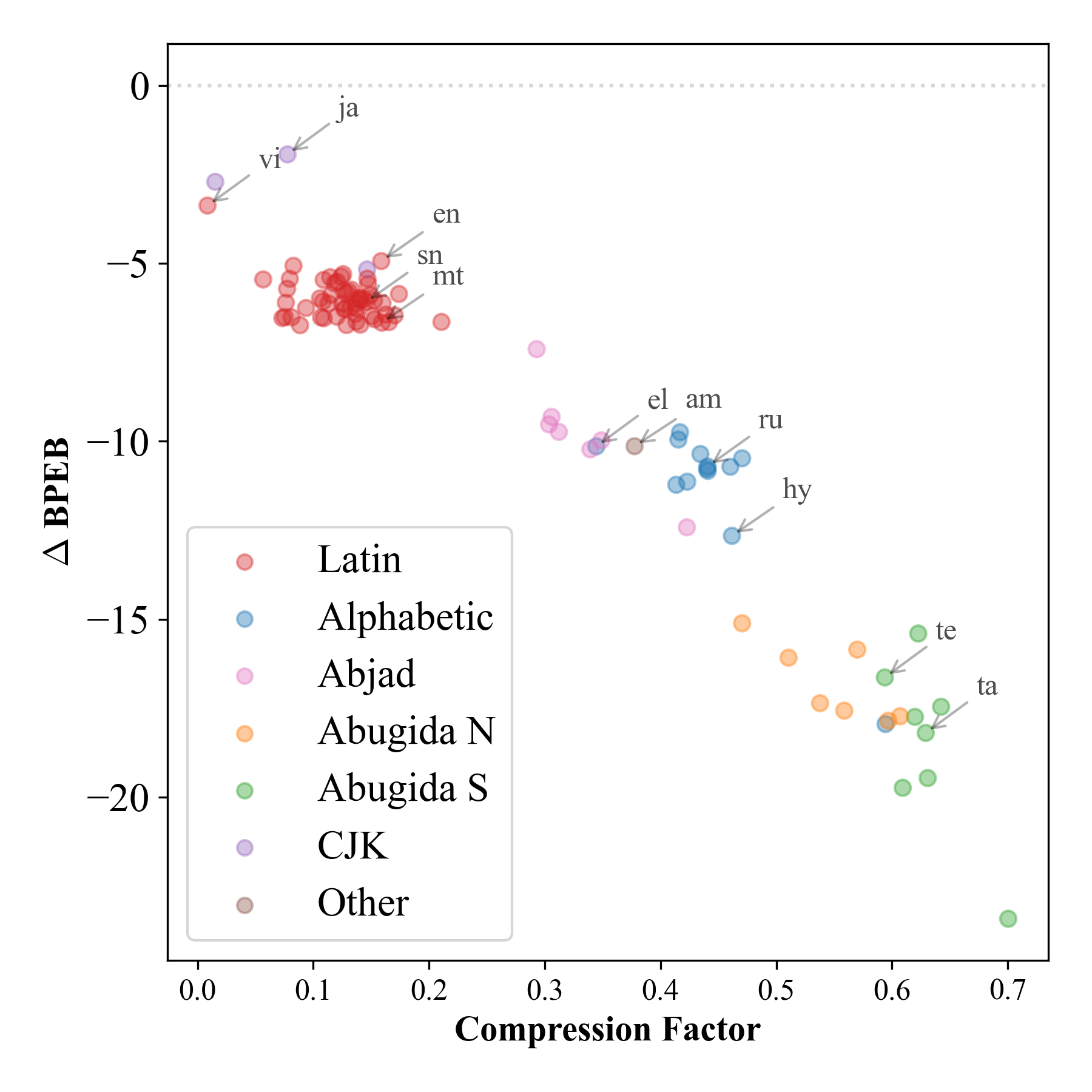}
        \caption{LM Performance ($\rho_S=-0.81$)}
        \label{fig:bpeb-delta}
    \end{subfigure}
    \hfill{}
    \begin{subfigure}[b]{0.44\linewidth}
        \includegraphics[width=\textwidth]{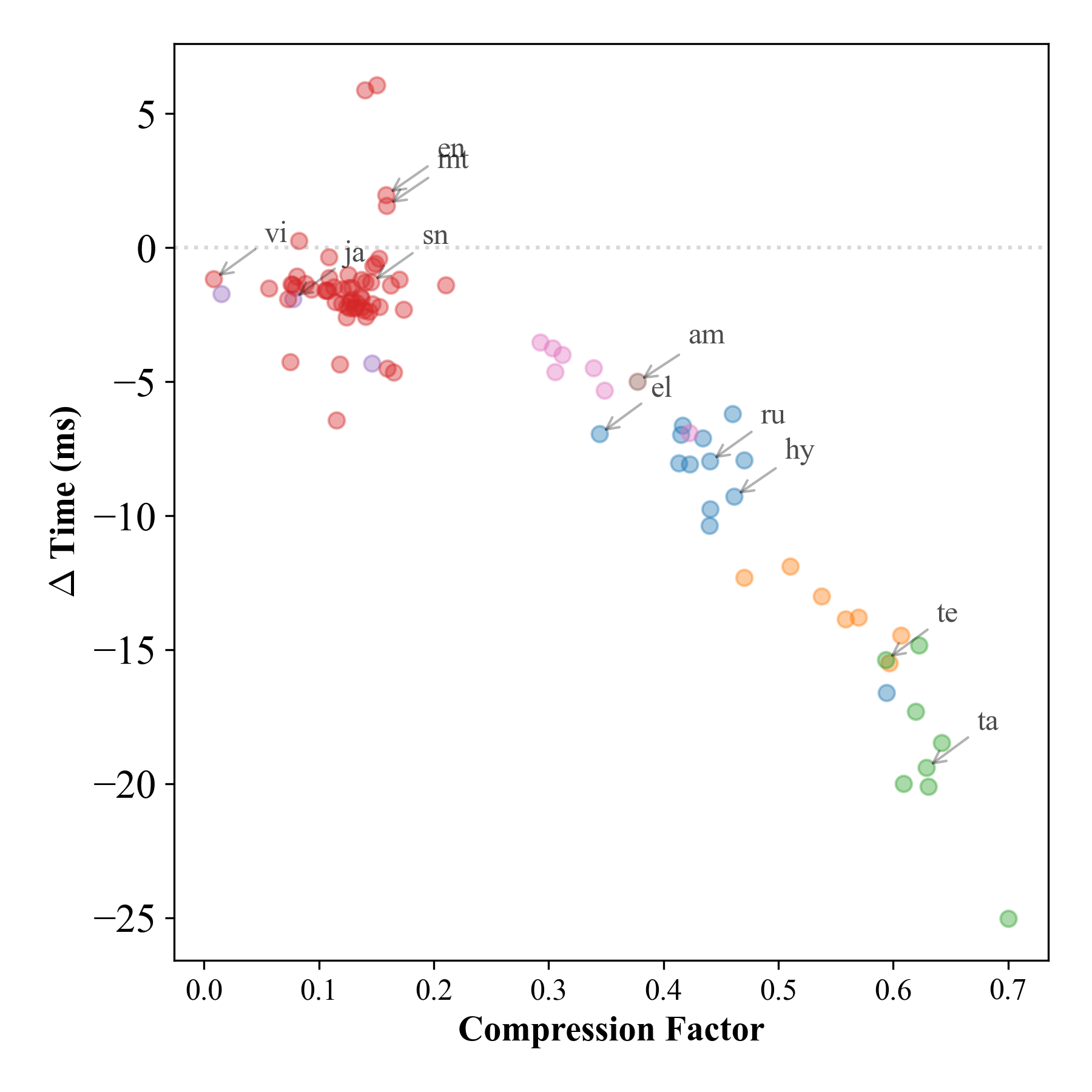}
        \caption{Inference Time ($\rho_S=-0.77$)}
        \label{fig:time-delta}
    \end{subfigure}
    \caption{The difference in Byte-per-English-Bit and inference time between \myt{} and \byt{} large models against compression factor of \ours{}. For each sentence, the BPEB value is normalized by the number of UTF-8 bytes used to represent the corresponding English sentence.
    The inference was run on A40 GPU core, we report an average per-sentence deltas.
    $\rho_S$ are Spearman's correlation coefficients.
    }
    
\end{figure*}

%% file: figures/lm_flores.tex
\begin{figure*}[!t]
    \centering   \includegraphics[width=\textwidth]{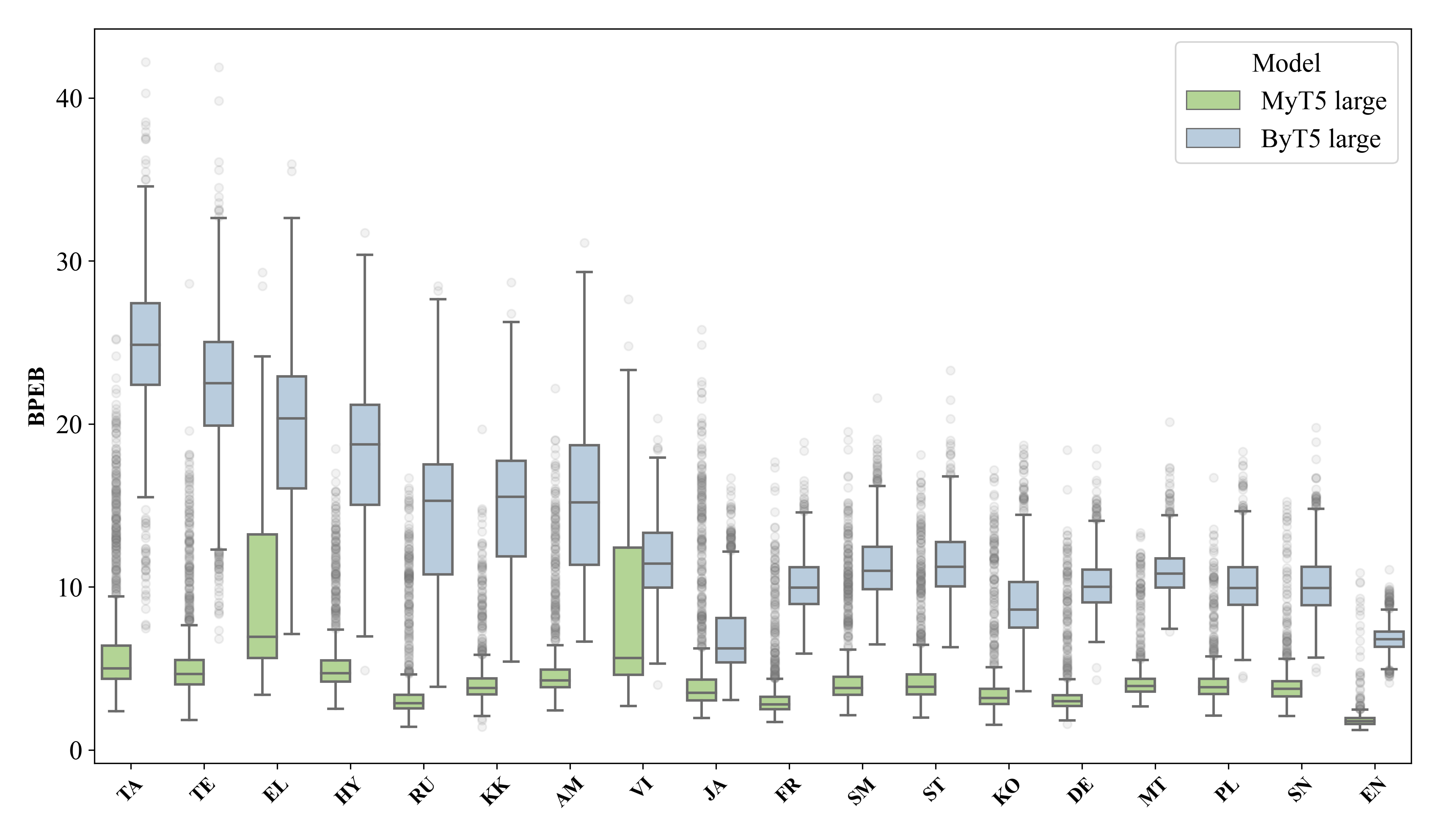}
    \caption{Sentence prediction suprisal expressed as Bit-per-English-Byte on multi-parallel Flores 200 corpus. 
    Each point corresponds to the BPEP value of one sentence. 
    The comparison shows that under \myt{} model, performance is more equitable across languages than in the standard \byt{} model.}
    \label{fig:flores-lm}
\end{figure*}

%% file: figures/lm_flores_unseen.tex
\begin{figure}[!t]
    \centering   
    \includegraphics[width=\linewidth]{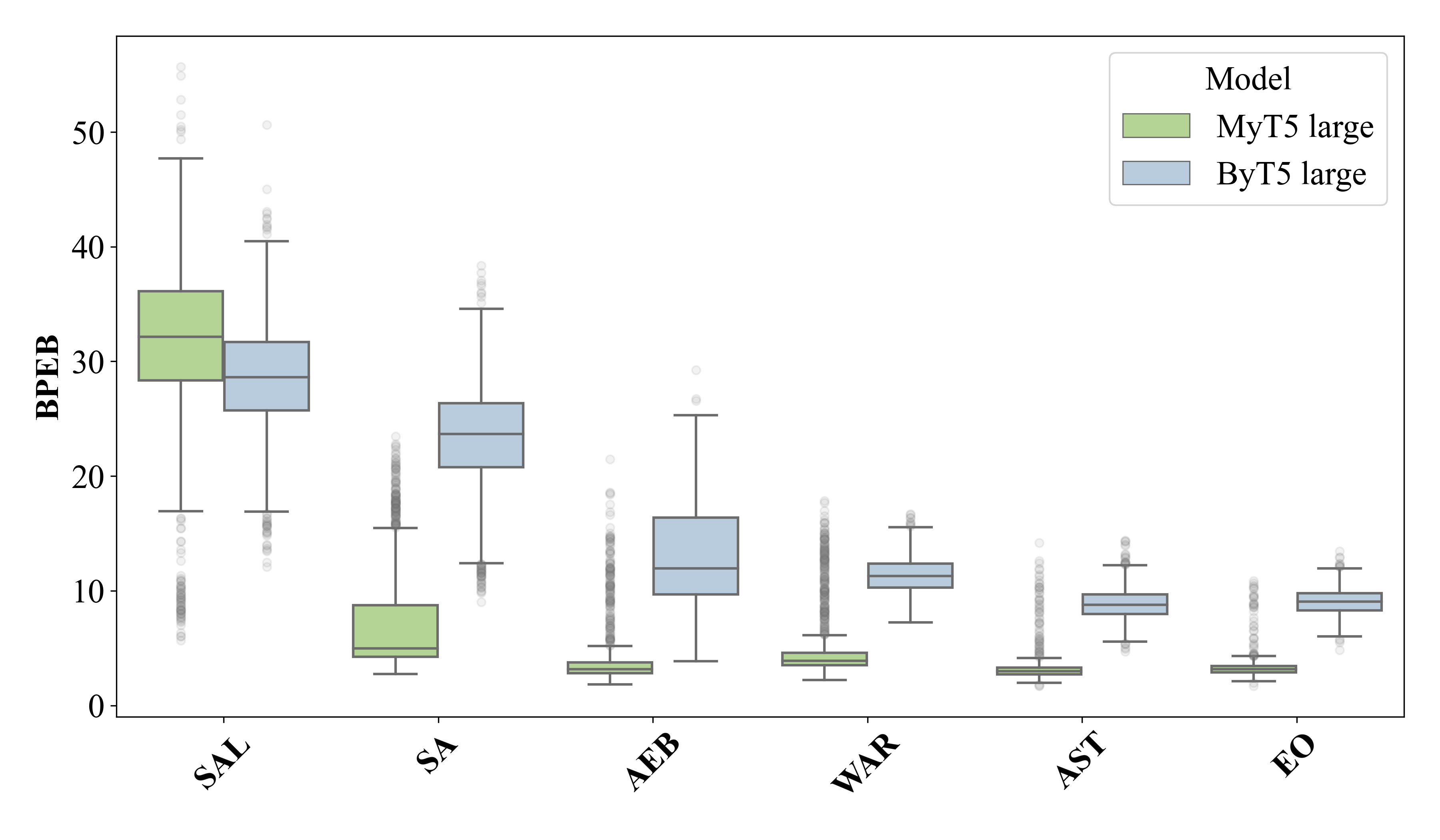}
    \caption{Bit-per-English-Byte for six languages unseen in morphological analysis: Santali, Sanskrit, Tunisian, Waray, Asturian, Esperanto. Santali (\emph{sal}) uses an unseen script (Ol Chicki).}
    \label{fig:flores-lm-unseen}
\end{figure}

%% file: tables/aggregated_modeling.tex
\begin{table}[!t]
\centering
\small
\begin{tabular}{llcccc}
\toprule
      &           & \multicolumn{2}{c}{Byt5} & \multicolumn{2}{c}{Myt5} \\\cmidrule(l){3-4} \cmidrule(l){5-6}  
      &           & BPEB & T (ms) & BPEB & T (ms) \\
\midrule
small & All & 10.1 &       7.0 &  4.6 &       6.7 \\
      & Latin &  4.6 &       5.9 &  4.2 &       6.6 \\
      & Non Latin & 18.1 &       8.5 &  5.1 &       6.8 \\ \midrule
base & All &  8.2 &      11.5 &  5.8 &       8.9 \\
      & Latin &  4.9 &       9.4 &  5.0 &       8.7 \\
      & Non Latin & 13.0 &      14.6 &  6.9 &       9.1 \\ \midrule
large & All & 13.4 &      31.8 &  4.6 &      26.7 \\ 
      & Latin & 10.1 &      28.1 &  4.0 &      26.6 \\
      & Non Latin & 18.2 &      37.3 &  5.4 &      27.0 \\
\bottomrule
\end{tabular}
\caption{Byte-per-English-Bits and Inference times (average per Flores 200 sentence) averaged for three language groupings. }
\label{tab:aggregated_modeling}
\end{table}

%% file: figures/xtreme_up_comparison.tex
\begin{figure}[!tb]
    \centering
    \includegraphics[width=\linewidth]{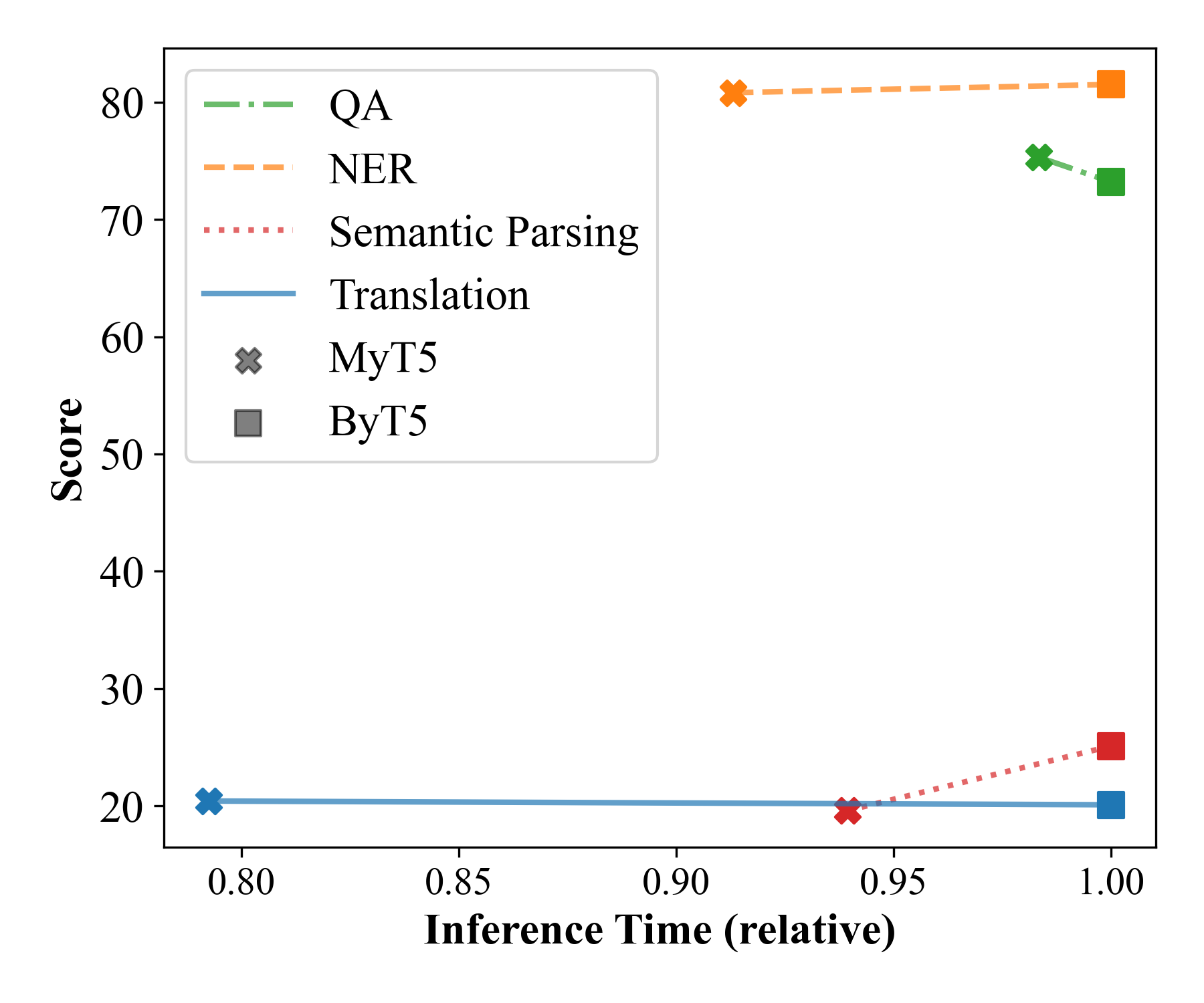}
    \caption{Avarage score on end tasks from \xu{} end tasks on low-resource languages against the inference time. 
    The times were divided by the value for \byt{} model, which is always higher than \myt{} model.
    The metrics and the absolute values of inference time are shown in Table~\ref{tab:xtreme_up}.}
    \label{fig:xtreme_up_comparison}
\end{figure}

%% file: tables/xtreme_up_all.tex
\begin{table}[!t]
\centering
\small
\begin{tabular}{lcccc}
\toprule
Task &   QA &  NER & \thead{Semantic \\ Parsing} & MT \\ \cmidrule(l){2-5}
Metric &   F1 &  F1 & EM & chrF \\ 
\midrule
Flan-PaLM* & 22.9 & 12.0 & 0.1 & --- \\ 
mT5* & 59.7 & 74.0 & 21.8 & --- \\
\midrule
ByT5 & 73.2 & 81.5 &   25.1 &   20.1 \\ 
MyT5 & 75.3 & 80.8 &   19.6 &    20.4 \\
\midrule
\multicolumn{5}{c}{Inference Time (ms)} \\
\midrule
ByT5  & 36.2 & 13.8 &   13.2 &        15.9 \\
MyT5  & 35.6 & 12.6 &   12.4 &        12.6 \\ 
\bottomrule
\end{tabular}
\caption{The average result of \xu{} tasks across low-resource languages.
The baseline results of mT5 and Flan-PaLM (in-context-learning evaluation) are copied from: \citet{ruder_xtreme-up_2023}.
We observed disparities between their reported and reimplemented ByT5 results, which are probably caused by the differences in fine-tuning setting.
The time is an average across evaluation examples, the inference was run on an A40 GPU core.
The results for all languages and fine-tuning details are in Appendix.}
\label{tab:xtreme_up}
\end{table}

%% file: sections/06_related_work.tex
\section{Related Work}

\subsection{Fair Representation across Languages}

Perhaps the most significant challenge of multilingual NLP is the large disparity of resourcefulness across the world's languages \cite{joshi-etal-2020-state}, as the size and quality of data used for the model training directly affects its performance in individual languages.
Hence, researchers have proposed multiple ways to balance the training signal across languages \cite{malkin-etal-2022-balanced}. 
Solutions include sampling data to overrepresent low-resource languages, e.g., with alpha \cite{conneau-etal-2020-unsupervised} or uniform sampling of data across languages \cite{chung2023unimax}. This unequal treatment of languages is also present in how data is encoded as input to the model \cite{ahia_all_2023}. 
\citet{petrov_language_2023} show that practically all methods used to represent texts as input of NLP systems treat languages unequally, segmenting some (mainly the lowest-resourced ones) into fine-grained non-informative units.

Some approaches aimed at balancing the segmentation or tokenization methods have been introduced.
\citet{limisiewicz-etal-2023-tokenization} proposed merging vocabulary based on the tokenizer scoring function.
\citet{zheng_allocating_2021} introduced a method of allocating vocabulary capacity uniformly across languages, while \citet{chung_improving_2020} constructed multilingual vocabulary for clusters of languages and merged them. 
\citet{liang-etal-2023-xlm} combined the elements of both approaches and showed the advantage of extending vocabulary to benefit multilingual transfer.
These solutions promise to obtain a better allocation of vocabulary units.
However, they do not solve the inequality of the underlying encoding, which may affect the construction process of vocabulary units.
For instance, byte merges in the BPE algorithm begin at individual bytes \citet{sennrich_neural_2016, zouhar-etal-2023-formal}.
Therefore, the unequal granularity of \utf{} representation impacts the vocabulary construction step in BPE, especially harming the low-resource non-Latin languages \cite{kargaran-etal-2024-glotscript-resource}.
A possible solution is training BPE on top of \ours{} encoded and balanced multilingual corpus.

Morphological analyzers, such as Morfessor, showed promising results for segmenting input texts for language models and neural machine translators \cite{Machcek2018MorphologicalAL,hou-etal-2023-effects}. We are the first to apply morphology-based encoding for a massively multilingual setting.



\subsection{Tokenization-free Language Modeling}

An alternative to subword tokenization is representing texts directly as underlying encoding: characters or bytes. 
Or even representing texts as pixels of rendered text images \cite{rust-etal-2023-pixel}.

\citet{xue_byt5_2022} shows that for many non-Latin scripts, byte-level encoding performs worse than subword tokenization.
The problem with small units is that they do not carry meaningful information independently and often underperform subword models \cite{sun-etal-2023-multi,clark-etal-2022-canine}.

The researchers have proposed multiple algorithms to enrich the byte-level embeddings with information from a local context.
For that purpose, recent approaches use shallow networks to aggregate information in local contexts defined as character n-grams \cite{clark-etal-2022-canine},  byte patches \cite{yu2023megabyte}, or character blocks \cite{tay2021charformer}. 
However, the problem with choosing the appropriate context window is hard, because information density varies for different languages.
A solution to that problem can be dynamically learning the segmentation in byte sequences \cite{nawrot-etal-2023-efficient}.
Another approach is to redefine the encoding convention to equate the information loads in sequences, as the proposed \ours{} approach.

%% file: sections/07_conclusion.tex
\section{Conclusion}

In this paper, we introduce \ours{} encoding, a fairer byte-level representation for multilingual language modeling that is based on morphological segmentation. We show that adapting a morphological analyzer to unsupervised segmentation allows us to represent multi-parallel corpora with comparable encoding lengths across a wide range of languages. Additionally, our new representation significantly improves language modeling, especially of low-resource and non-Latin script languages, and provides efficiency benefits over traditional byte-level models. These trends hold across model sizes, with improvement increasing at scale. 
Overall, \ours{} bridges the gap in encoding efficiency between high and low-resource languages, benefiting (to varying extent) all \emph{99} analyzed languages.
\clearpage

%% file: sections/08_es_limitations.tex
\section*{Ethical Statement}
Our work makes a significant contribution to a fairer representation of text across diverse languages.
It will potentially benefit the speakers of underrepresented languages by enabling access to more reliable and cheaper NLP tools.
For all the experiments, we relied on open-source tools and datasets.
We strongly discourage unintended usage of the released language models.

\section*{Limitations}

Our method inherits the limitations of Morfessor, which was used to obtain multilingual morphological segmentation for \ours{}.
First, Morfessor is data dependent and is affected by the quality of the corpus (Wikipedia) and the lexicon (MUSE \citet{conneau_word_2017} when available). The artifact of these resources is a significant presence of cross-lingual contamination, typically from high-resource languages \cite{blevins-zettlemoyer-2022-language}. This leads to the appearance of Latin (typically English) morphemes in analyses of many languages.
Second, we use the unsupervised mode of Morfessor that can be applied to any language due to its independence of annotated data. However, it is also prone to errors in morphological segmentations, i.e., oversegmenting texts of specific languages. We mitigate this issue by picking a constant target number of morphemes.

Dependence on data might also affect the generalizability of our findings' to the languages that were not used in the construction of \ours{}. Results in Section \ref{sec:results-sequences} show that the method is indeed effective in compressing text representation of unseen languages but not unseen scripts.
Notably, we do not exhaust the capacity of the \ours{} codepage; thus, it can be extended to further languages. 

Lastly, even perfect morphological analysis cannot guarantee equal granularity of segmentation across languages. Some languages are characterized by higher morphological richness, thus their texts consist of more morphemes. Accordingly, we observe differences in \ours{} segmentation lengths across languages, yet these disparities are significantly smaller than in other conventions.

%% file: sections/91_morphological_analysis.tex
\section{Details of Unsupervised Morphological Analysis}
\label{sec:app-morphology}

In this appendix, we provide details on the prerequisites of \ours{} transcoding algorithm: a) preparing multilingual lexicons and corpora for morphological analysis and b) usage of Morfessor unsupervised algorithm to obtain morpheme inventory for each language.

\subsection{Preparing Lexicons for Morphological Analysis}

To obtain morphological segmentation across a wide variety of languages and scripts, we perform the following steps:

\begin{enumerate}
    \item We use 45 languages with bilingual lexicons available through MUSE \cite{conneau_word_2017} as a base.
    Lexicons are obtained independently for each language; hence, we ignore the bilingual aspect of the data.
    We filter out the lexemes that are the same in English and the target language to avoid contamination that would unfairly boost the frequency of English words across lexicons.
    \item We use Wikipedia corpus dump from September 2023 \url{dumps.wikimedia.org} to count the occurances of lexemes. 
    For 54 languages included in mC4 \cite{raffel_exploring_2019}, but without the MUSE lexicon, we compile the list of unique words in Wikipedia as a lexicon.
    \item The lexicons are clipped to the size of 30,000 lexemes. 
    \item All lexemes are transcribed to bytes via \utf{} standard. 
    All byte sequences are decomposed following NKFD convention, i.e., modifying symbols (diacritics, accents), which are represented as separate codepoints. 
    On top of \utf{} decomposition, we rewrite capital letter codes into lowercase letters and capitalization markers. 
\end{enumerate}

\subsection{Unsupervised Segmentation with Morfessor}


We use Morfessor \cite{smit-etal-2014-morfessor}, an unsupervised algorithm producing segmentation on a sub-word level that resembles morphological analysis.
The unsupervised nature of the method allows us to apply it to a wide range of languages.
However, it is essential to note that the method is prone to errors, such as over-segmentation of roots or misplaced morpheme boundaries.
We use adaptive loss weighting to limit the number of attested morphs to around 4096 to avoid over-segmentation. 
Unlike the typical usage of Morfessor, we applied it to the corpus on byte instead of character level.

\subsection{Morfessor: Technical Details}

Morfessor uses recursive optimization to produce subword segmentation akin to morphological analysis.
The input data required for unsupervised analysis are language corpus and lexicon consisting of unique words $c \in \mathcal{C}$. We also define the set of atoms $a \in \mathcal{A}$, which are indivisible segments of texts that can be assembled into words. We choose atoms to be \utf{} bytes.

The aim of the algorithm is to find a set of morphemes $m \in \mathcal{M}$ appearing in the segmentation of words from the given lexicon. The set of morphemes $\mathcal{M}$ is extended by a recursive algorithm optimizing two losses: \emph{corpus loss} and \emph{lexicon loss} computed with respective data resources. 
Before providing equations for the mentioned losses, let's define the auxiliary variables:

\begin{equation}
\begin{split}
    M = \sum_{m \in \mathcal{M}} \#_{\text{COR}}(m) \\
    C = \sum_{c \in \mathcal{C}} \#_{\text{COR}}(m) -1 \\
    A = \sum_{a \in \mathcal{A}} \#_{\mathcal{M}}(a)
\end{split}
\end{equation}

The $\#$ notation is used to denote the number of elements in the corpus (COR) or morpheme set $\mathcal{M}$.
In other words, $M$ is the total number of morphemes in the corpus, $C$ is the total number of words in the corpus, and $A$ is the total number of atoms in the set of (unique) morphemes. Morfessor uses the following losses in recursive optimization:

\paragraph{Corpus loss} favors morphemes frequently appearing in the corpus:

\begin{equation}
\begin{split}
   \mathcal{L}_{\text{COR}} = (M+C)\log(M+C) + \\
    - \sum_{m \in \mathcal{M}} \#_{\text{COR}}(m)\log\#_{\text{COR}}(m) + \\
    + \log\binom{M - 1}{|\mathcal{M}| - 1}
\end{split}
\end{equation}

\paragraph{Lexocon loss} favors segments consisting of diverse sets of atoms so that overlapping segments are not identified as morphemes:

\begin{equation}
\begin{split}
   \mathcal{L}_{\text{LEX}} = (A + |\mathcal{M}|)\log(A + |\mathcal{M}|)
   - |\mathcal{M}|\log|\mathcal{M}| + \\
   - \sum_{a \in \mathcal{A}} \#_{\mathcal{M}}(a)\log\#_{\mathcal{M}}(a) - log(|\mathcal{M}|!) + \\
   + \log\binom{A - 1}{|\mathcal{A}| - 1} 
\end{split}
\end{equation}

The losses are weighted by a parameter $\alpha$, which indirectly controls the size of the morpheme set $|\mathcal{M}|$. For instance, we adapt $\alpha$ to keep the number of morphemes close to 4096 for each language. We observed that this size leads to comparable segmentation across languages,

\begin{equation}
\mathcal{L} = \alpha\mathcal{L}_{\text{COR}} + \mathcal{L}_{\text{LEX}} 
\end{equation}

%% file: sections/92_supplementary_results.tex
\section{Supplementary Results}
\label{sec:app-results}

\input{figures/unseen_byte_sequences_comparison}
\input{figures/bseq_lengths}

\input{figures/compression_rates_ordered}

\input{figures/bpeb_delta_small_base}

\input{tables/results_per_language}

This appendix summarizes complementary results referred to 
throughout the papers. 

\subsection{Results for Each Language}
All the experimental results for each of the analyzed mC4 languages are presented in Table~\ref{tab:results_per_language}.
Sequence lengths under \utf{} and \ours{} are visualized in Figure~\ref{fig:bseq-len}.
Corresponding compression rates in Figure~\ref{fig:comp-rates-ord}.

Figure~\ref{fig:bseq-unseen-comparison} illustrates the sequence lengths and compressions obtained for languages \emph{unseen} in the morphological analysis. While Figure~\ref{fig:flores-lm-unseen} shows the comparison of \byt{} and \myt{} for these languages.

\subsection{LM Performance Across Scales}
Figure~\ref{fig:bpeb-delta-small-base} shows the difference of $BPEB$ between \myt{} and \byt{} the models in small and base scales. 
Furthermore, Table~\ref{tab:results_per_language} contains average language modeling scores and inference times across all available scales for each language.
Both Figures show that \ours{} offers improvement for languages not seen in morphological analysis but not for Sanatli, which uses a distinct script. 

\input{tables/xtreme_up_qa}
\input{tables/xtreme_up_ner}
\input{tables/xtreme_up_sp}
\input{tables/xtreme_up_mt}

\subsection{\xu{} Benchmark Results}
Tables \ref{tab:xtreme_up_qa}, \ref{tab:xtreme_up_ner}, \ref{tab:xtreme_up_sp}, and \ref{tab:xtreme_up_mt} present detailed results of edtasks collected in \xu{} benchmark.

%% file: figures/unseen_byte_sequences_comparison.tex
\begin{figure}[!tb]
    \centering
    \begin{subfigure}[b]{\linewidth}
        \includegraphics[width=\textwidth]{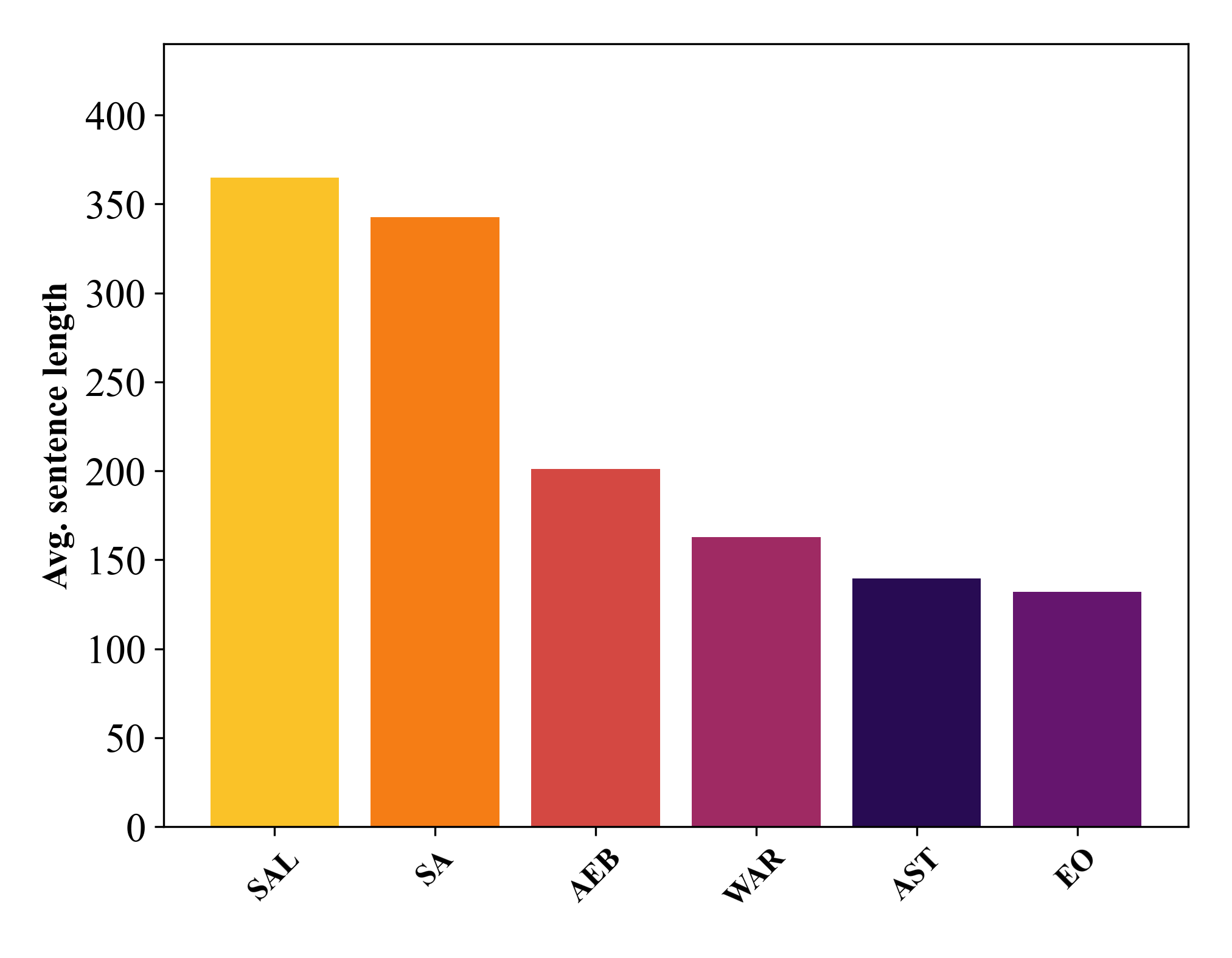}
        \caption{\utf{}}
    \end{subfigure}
    \begin{subfigure}[b]{\linewidth}
        \includegraphics[width=\textwidth]{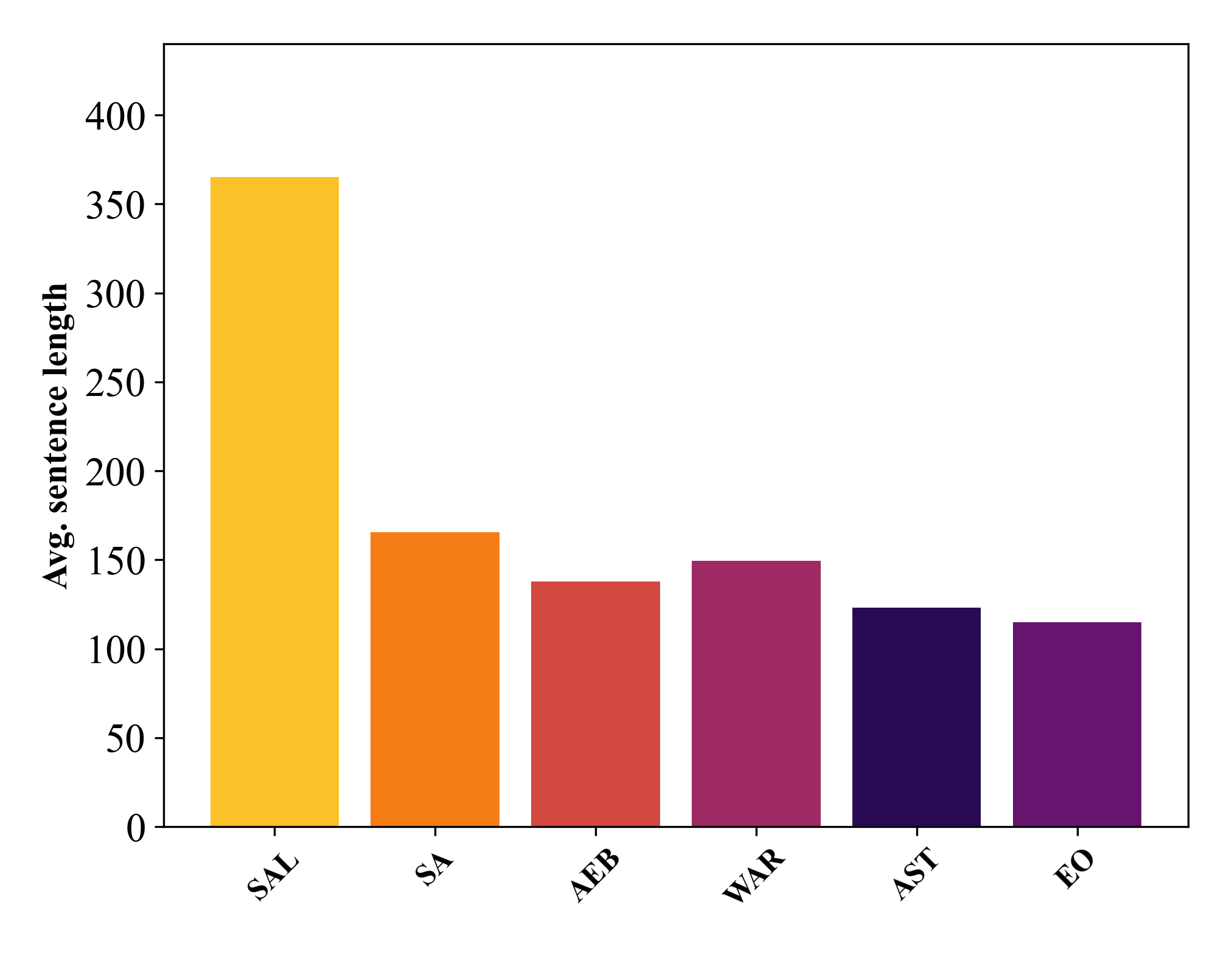}
        \caption{\ours{}}
    \end{subfigure}
    \begin{subfigure}[b]{\linewidth}
        \includegraphics[width=\textwidth]{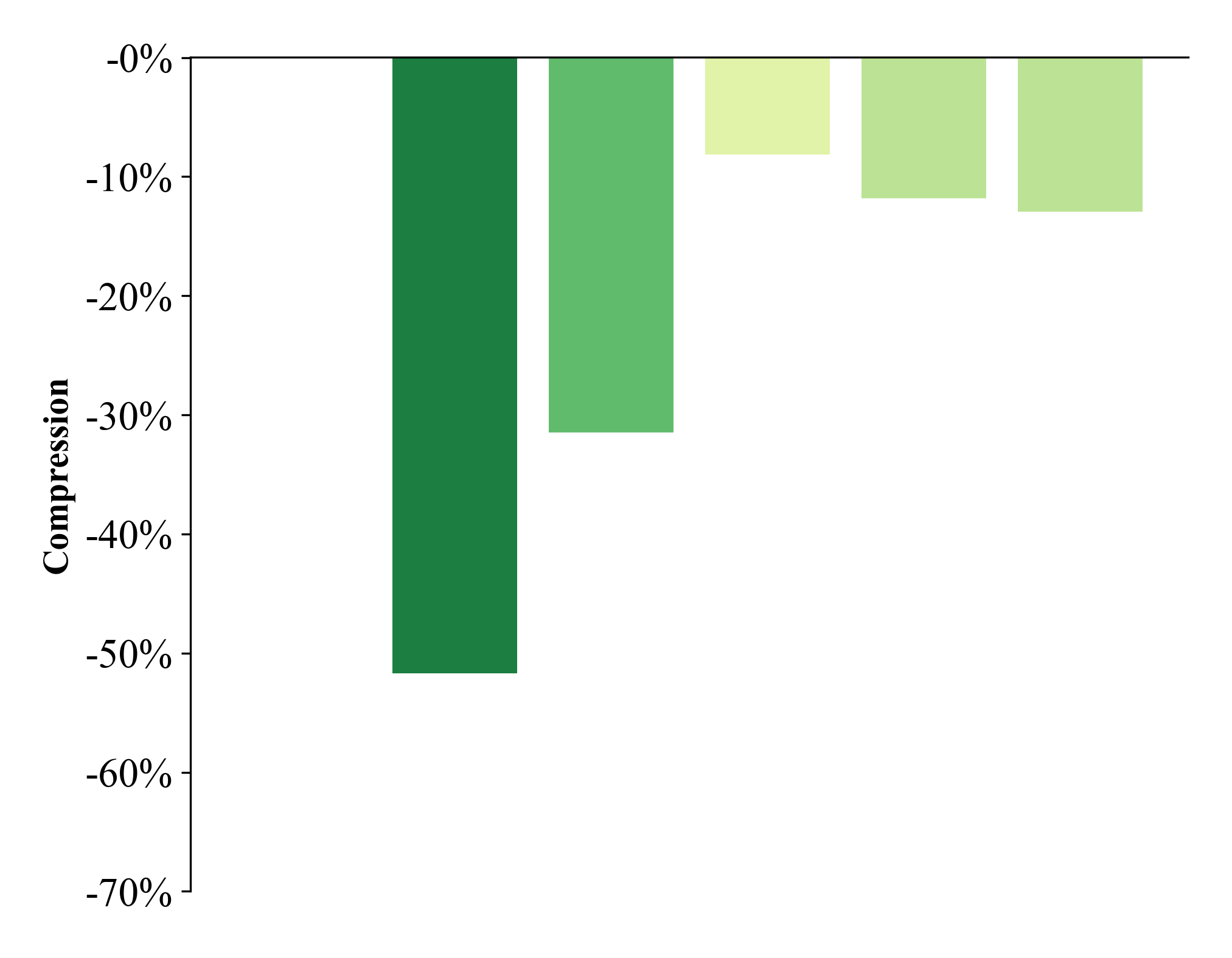}
        \caption{Sequence compression}
    \end{subfigure}
    \caption{Average byte sequence lengths of parallel sentences for languages unseen in the morphological analysis used in the construction of \ours{}. Santali (\emph{sal}) uses a script (Ol Chicki), a distinct script not seen in morphological analysis. }
    \label{fig:bseq-unseen-comparison}
\end{figure}

%% file: figures/bseq_lengths.tex
\begin{figure*}[!tb]
    \centering
    \begin{subfigure}[b]{\textwidth}
        \includegraphics[width=\textwidth]{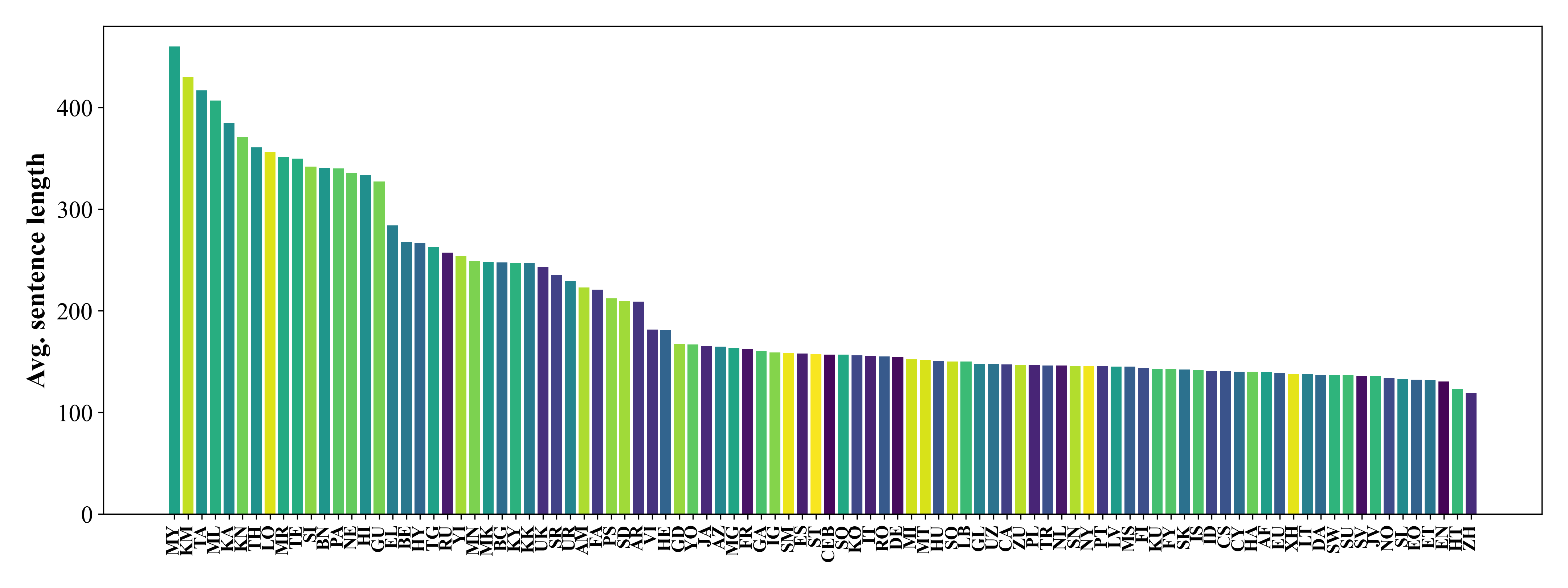}
        \caption{Original UTF-8}
    \end{subfigure}
    \begin{subfigure}[b]{\textwidth}
        \includegraphics[width=\textwidth]{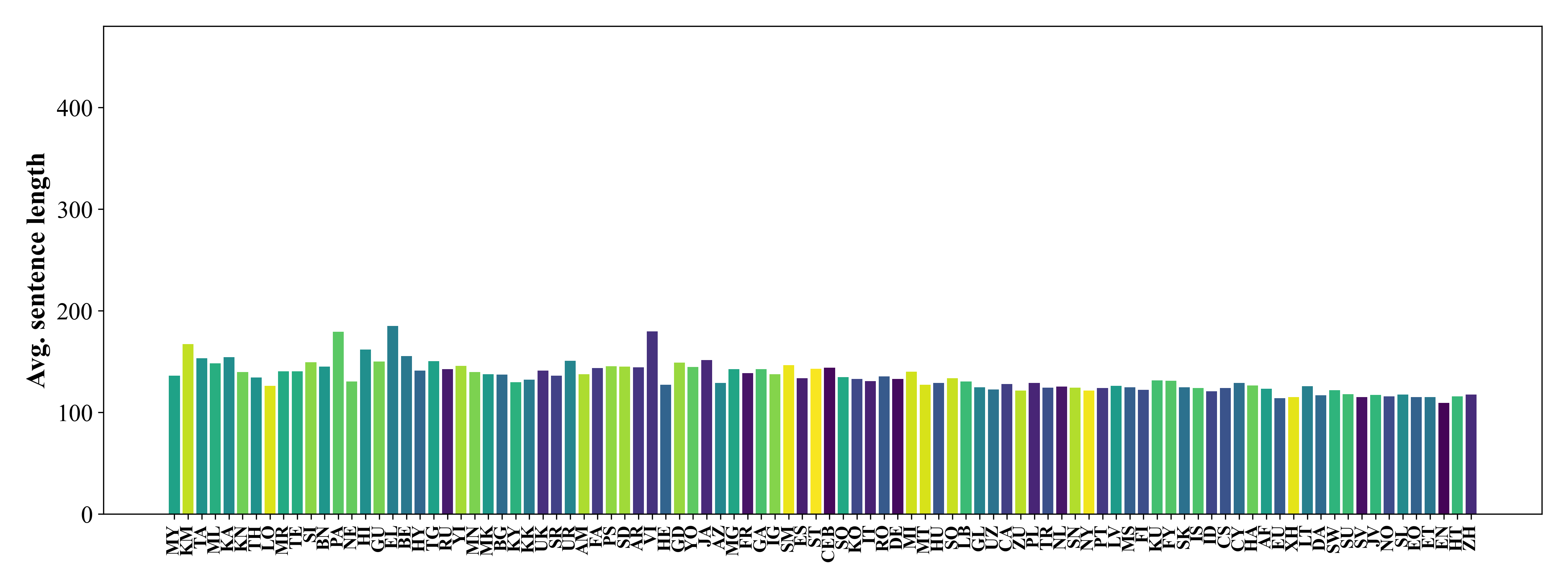}
        \caption{\ours{} (Same order as in a)}
    \end{subfigure}
    \caption{Average byte sequence lengths of parallel sentences from Flores 200.}
    \label{fig:bseq-len}
\end{figure*}

%% file: figures/compression_rates_ordered.tex
\begin{figure*}[!tb]
    \centering
    \includegraphics[width=\textwidth]{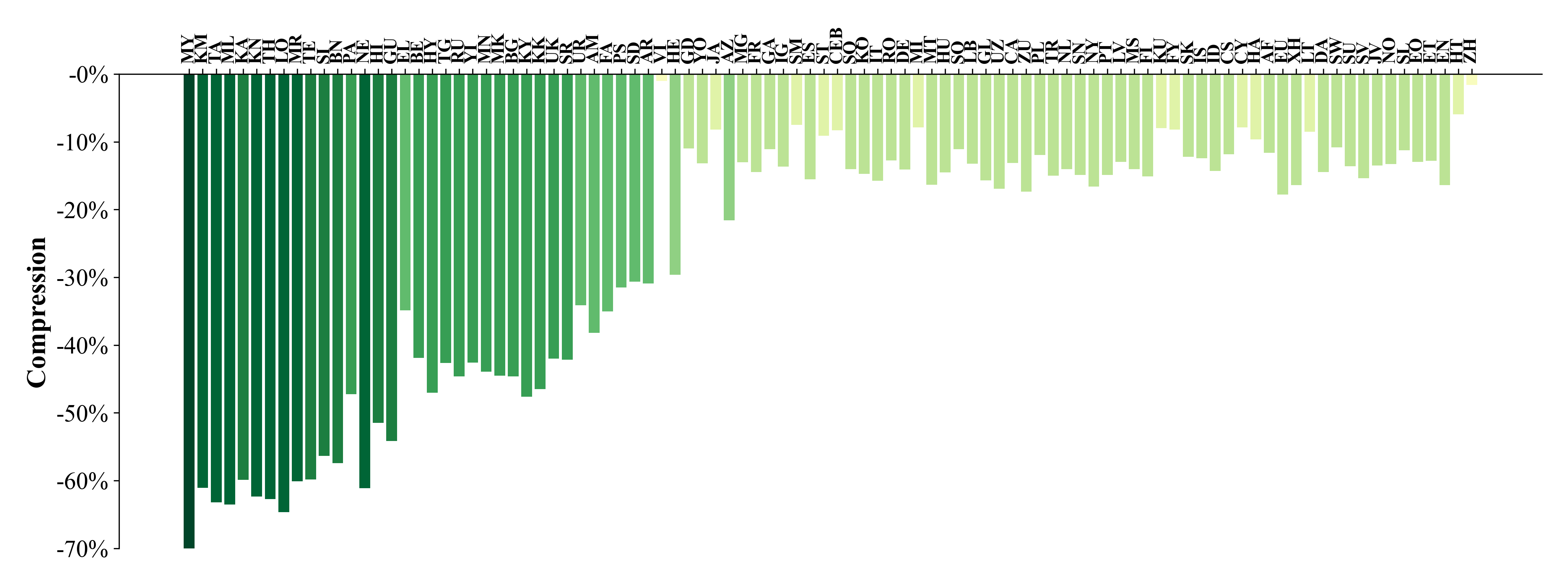}
    \caption{Sequence compression rates on Flores 200 of \ours{} in comparison with the original \utf{} encoding.}
    \label{fig:comp-rates-ord}
\end{figure*}

%% file: figures/bpeb_delta_small_base.tex
\begin{figure*}[!tb]
\centering
    \begin{subfigure}[b]{0.48\linewidth}
        \includegraphics[width=\textwidth]{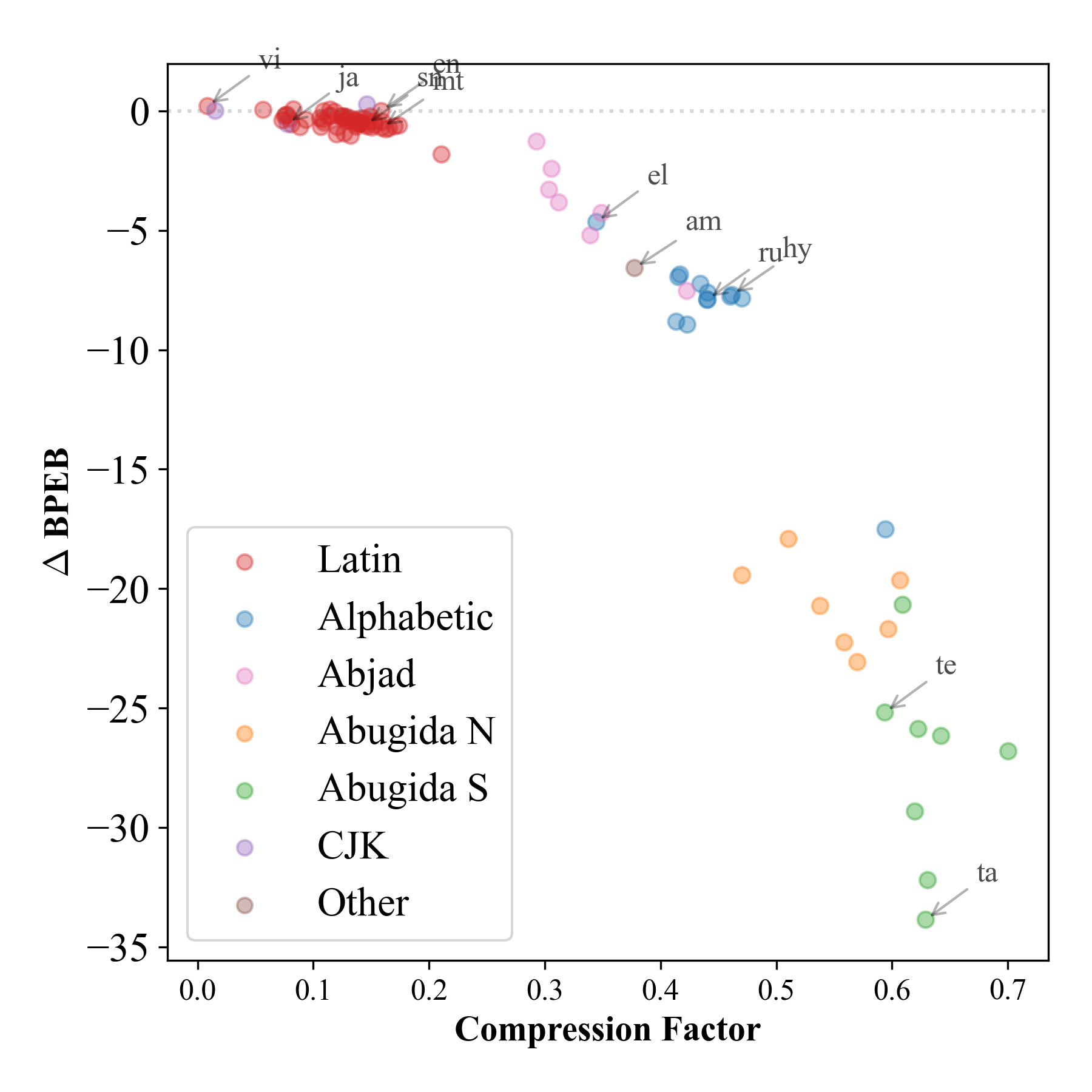}
        \caption{small}
    \end{subfigure}
    \hfill{}
    \begin{subfigure}[b]{0.48\linewidth}
        \includegraphics[width=\textwidth]{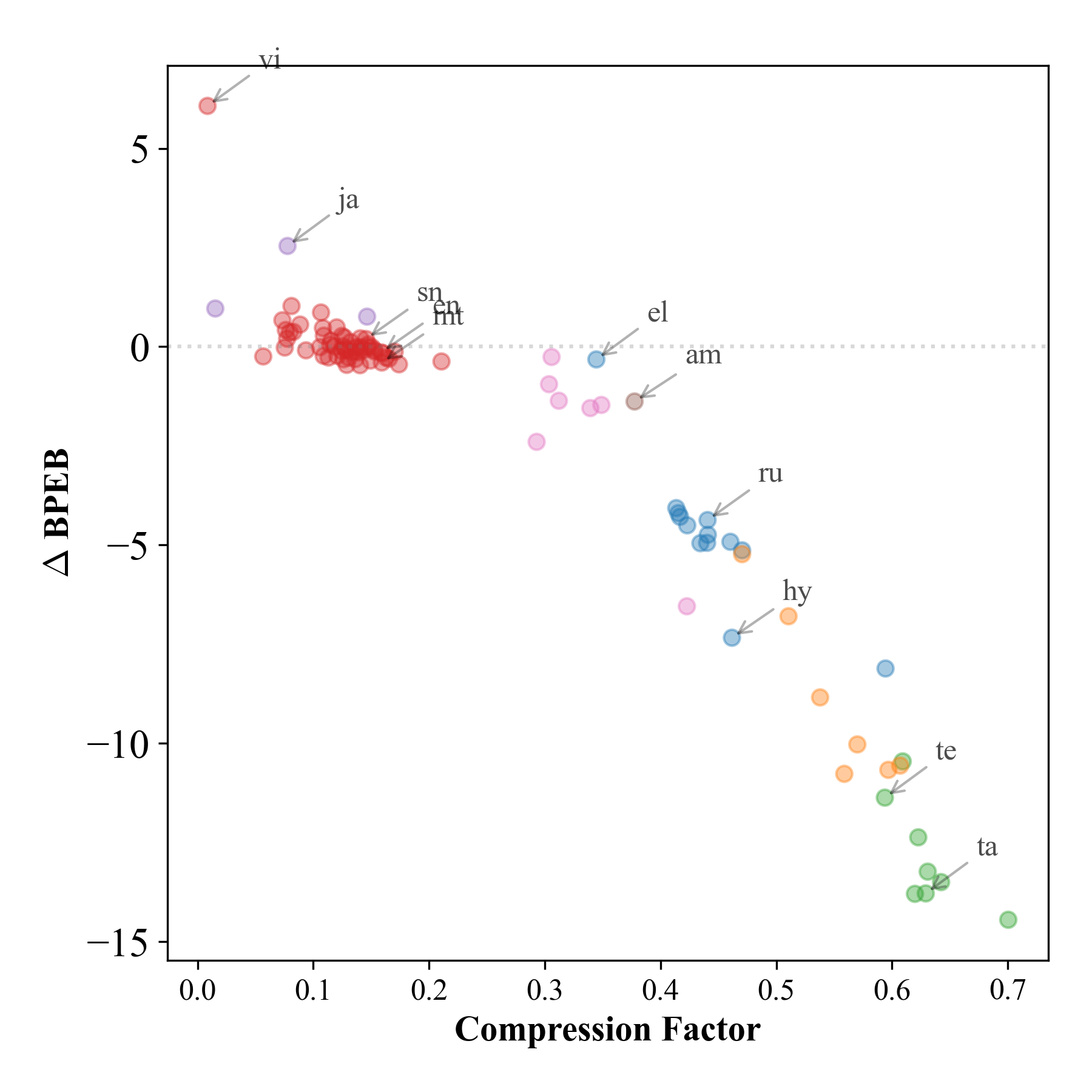}
        \caption{base}
    \end{subfigure}
    \caption{The difference in Byte-per-English-Bit between \myt{} and \byt{} for models in small and base scales.
    }
    
    \label{fig:bpeb-delta-small-base}
\end{figure*}

%% file: tables/results_per_language.tex
\afterpage{
\thispagestyle{empty}
\begin{table*}[!tb]
\centering
\tiny
\begin{tabular}{cccccccccccccccccc}
\toprule
\multirow{3}{*}{Lang} & \multicolumn{2}{c}{\utf{}} & \multicolumn{2}{c}{\ours{}} & \multirow{2}{*}{Comp.} . & \multicolumn{6}{c}{ByT5} & \multicolumn{6}{c}{MyT5} \\ \cmidrule(r){2-3} \cmidrule(l){4-5} \cmidrule(r){7-12} \cmidrule(l){13-18}
& Parity & Len. & Parity & Len. & & \multicolumn{3}{c}{BPEB} & \multicolumn{3}{c}{Time (ms)} & \multicolumn{3}{c}{BPEB} & \multicolumn{3}{c}{Time (ms)}\\ \cmidrule(lr){7-9} \cmidrule(lr){10-12} \cmidrule(lr){13-15} \cmidrule(lr){16-18}
 &   &   & &   & in \% &  small &  base &  large &  small &  base &  large  &  small &  base &  large &  small &  base &  large \\
\midrule
af       &          1.1 &        139.6 &          1.1 &        123.3 &          11.7 &                    3.9 &                   4.6 &                    9.5 &                         5.8 &                        8.9 &                        27.6 &                    3.7 &                   4.3 &                    3.4 &                         6.6 &                        8.4 &                        26.1 \\
am       &          1.7 &        222.8 &          1.3 &        137.6 &          38.2 &                   11.7 &                   8.3 &                   15.3 &                         6.9 &                       12.5 &                        32.0 &                    5.1 &                   6.9 &                    5.2 &                         6.7 &                        9.0 &                        27.0 \\
ar       &          1.6 &        208.8 &          1.3 &        144.2 &          30.9 &                    7.0 &                   6.7 &                   13.7 &                         7.0 &                       11.6 &                        31.4 &                    4.6 &                   6.4 &                    4.4 &                         6.4 &                        9.1 &                        26.7 \\
az       &          1.3 &        164.6 &          1.2 &        129.1 &          21.6 &                    6.4 &                   5.7 &                   11.4 &                         6.3 &                        9.9 &                        28.9 &                    4.6 &                   5.4 &                    4.7 &                         6.7 &                        8.7 &                        27.5 \\
be       &          2.1 &        267.7 &          1.4 &        155.5 &          41.9 &                   14.6 &                  11.6 &                   17.1 &                         8.1 &                       13.5 &                        35.5 &                    5.7 &                   7.5 &                    5.9 &                         6.9 &                        9.5 &                        27.4 \\
bg       &          1.9 &        247.6 &          1.3 &        137.1 &          44.6 &                   11.9 &                   9.6 &                   14.6 &                         7.7 &                       12.6 &                        33.9 &                    4.3 &                   4.9 &                    3.8 &                         7.1 &                        8.8 &                        24.1 \\
bn       &          2.6 &        340.6 &          1.3 &        145.0 &          57.4 &                   28.4 &                  17.2 &                   21.3 &                         9.3 &                       16.6 &                        41.3 &                    5.3 &                   7.2 &                    5.4 &                         6.9 &                        9.2 &                        27.5 \\
ca       &          1.1 &        147.1 &          1.2 &        127.7 &          13.2 &                    4.1 &                   4.6 &                    9.6 &                         5.8 &                        9.4 &                        27.9 &                    3.9 &                   4.6 &                    3.4 &                         6.7 &                        8.5 &                        25.9 \\
ceb      &          1.2 &        156.9 &          1.3 &        143.8 &           8.3 &                    5.0 &                   5.0 &                   11.1 &                         6.0 &                        9.6 &                        28.7 &                    4.4 &                   6.0 &                    4.5 &                         6.3 &                        9.2 &                        27.6 \\
cs       &          1.1 &        140.7 &          1.1 &        124.1 &          11.8 &                    4.4 &                   4.8 &                    9.5 &                         5.5 &                        9.0 &                        28.1 &                    4.5 &                   4.8 &                    4.1 &                         6.7 &                        8.4 &                        26.0 \\
cy       &          1.1 &        140.1 &          1.2 &        129.1 &           7.9 &                    4.5 &                   5.1 &                   10.7 &                         5.9 &                        8.7 &                        27.6 &                    4.2 &                   5.1 &                    4.2 &                         7.1 &                        8.5 &                        23.4 \\
da       &          1.0 &        136.7 &          1.1 &        116.9 &          14.5 &                    3.9 &                   4.5 &                    9.3 &                         5.9 &                        8.7 &                        27.8 &                    3.6 &                   4.0 &                    3.3 &                         6.8 &                        8.2 &                        33.7 \\
de       &          1.2 &        154.5 &          1.2 &        132.8 &          14.1 &                    4.6 &                   5.1 &                   10.2 &                         6.2 &                        9.7 &                        27.9 &                    4.2 &                   5.1 &                    3.6 &                         6.5 &                        8.8 &                        26.7 \\
el       &          2.2 &        284.1 &          1.7 &        185.0 &          34.9 &                   12.8 &                  13.6 &                   19.5 &                         8.3 &                       14.0 &                        36.6 &                    8.2 &                  13.3 &                    9.3 &                         7.1 &                       10.4 &                        29.7 \\
en       &          1.0 &        130.5 &          1.0 &        109.1 &          16.4 &                    2.6 &                   3.3 &                    6.8 &                         6.2 &                       11.0 &                        28.4 &                    2.6 &                   3.2 &                    1.9 &                         6.6 &                       10.5 &                        30.3 \\
eo       &          1.0 &        132.2 &          1.1 &        115.1 &          12.9 &                    3.8 &                   4.4 &                    9.1 &                         5.8 &                        8.5 &                        27.2 &                    3.6 &                   4.1 &                    3.3 &                         6.6 &                        8.1 &                        25.7 \\
es       &          1.2 &        158.0 &          1.2 &        133.5 &          15.5 &                    4.7 &                   4.8 &                    9.9 &                         6.2 &                        9.2 &                        28.2 &                    4.0 &                   4.8 &                    3.4 &                         6.4 &                        8.7 &                        34.2 \\
et       &          1.0 &        131.9 &          1.1 &        115.0 &          12.8 &                    4.1 &                   4.4 &                    9.1 &                         5.8 &                        8.6 &                        27.3 &                    3.9 &                   4.3 &                    3.7 &                         6.4 &                        8.2 &                        24.7 \\
eu       &          1.1 &        138.6 &          1.0 &        114.0 &          17.8 &                    4.3 &                   4.6 &                    9.4 &                         5.9 &                        8.8 &                        27.9 &                    3.7 &                   4.1 &                    3.5 &                         6.5 &                        8.1 &                        25.6 \\
fa       &          1.7 &        220.9 &          1.3 &        143.4 &          35.1 &                    8.6 &                   7.6 &                   14.5 &                         8.7 &                       12.0 &                        32.0 &                    4.4 &                   6.1 &                    4.5 &                         7.0 &                        9.0 &                        26.7 \\
fi       &          1.1 &        144.1 &          1.1 &        122.3 &          15.1 &                    4.7 &                   4.8 &                    9.7 &                         5.8 &                        9.4 &                        26.6 &                    4.2 &                   4.9 &                    4.1 &                         6.7 &                        8.4 &                        25.9 \\
fr       &          1.2 &        162.1 &          1.3 &        138.6 &          14.5 &                    4.7 &                   5.1 &                   10.2 &                         6.3 &                        9.6 &                        28.2 &                    4.2 &                   5.3 &                    3.5 &                         6.5 &                        9.0 &                        26.9 \\
fy       &          1.1 &        143.0 &          1.2 &        131.3 &           8.2 &                    4.8 &                   5.1 &                    9.8 &                         5.7 &                        9.1 &                        27.6 &                    4.5 &                   5.5 &                    4.4 &                         6.5 &                        8.8 &                        26.1 \\
ga       &          1.2 &        160.3 &          1.3 &        142.5 &          11.1 &                    5.4 &                   5.7 &                   11.2 &                         6.0 &                        9.7 &                        28.1 &                    5.0 &                   6.2 &                    5.1 &                         6.7 &                        9.1 &                        26.5 \\
gd       &          1.3 &        167.2 &          1.4 &        148.8 &          11.0 &                    5.8 &                   6.0 &                   12.0 &                         6.1 &                       10.5 &                        28.9 &                    5.1 &                   6.9 &                    5.5 &                         6.8 &                        9.1 &                        27.3 \\
gl       &          1.1 &        148.0 &          1.1 &        124.8 &          15.7 &                    4.2 &                   4.5 &                    9.4 &                         6.1 &                        9.4 &                        28.0 &                    3.8 &                   4.4 &                    3.3 &                         6.7 &                        8.5 &                        27.6 \\
gu       &          2.5 &        327.1 &          1.4 &        150.0 &          54.1 &                   26.5 &                  16.6 &                   23.5 &                         8.9 &                       16.7 &                        39.8 &                    5.8 &                   7.8 &                    6.2 &                         6.8 &                        9.4 &                        26.8 \\
ha       &          1.1 &        140.0 &          1.2 &        126.5 &           9.6 &                    4.1 &                   4.6 &                    9.9 &                         5.7 &                        9.3 &                        27.5 &                    3.7 &                   4.5 &                    3.7 &                         6.5 &                        8.6 &                        25.9 \\
he       &          1.4 &        180.9 &          1.2 &        127.3 &          29.6 &                    5.6 &                   7.4 &                   11.3 &                         6.5 &                       10.2 &                        29.6 &                    4.3 &                   5.0 &                    3.9 &                         6.7 &                        8.5 &                        26.0 \\
hi       &          2.6 &        333.1 &          1.5 &        161.6 &          51.5 &                   23.8 &                  15.8 &                   22.2 &                         9.2 &                       16.2 &                        40.4 &                    5.9 &                   9.0 &                    6.1 &                         7.2 &                        9.7 &                        28.5 \\
ht       &          0.9 &        123.2 &          1.1 &        115.8 &           6.0 &                    3.5 &                   4.2 &                    8.7 &                         5.4 &                        8.4 &                        27.1 &                    3.5 &                   4.0 &                    3.3 &                         6.3 &                        8.0 &                        25.6 \\
hu       &          1.2 &        150.8 &          1.2 &        128.9 &          14.5 &                    5.2 &                   5.3 &                   10.5 &                         5.8 &                        9.5 &                        28.4 &                    4.7 &                   5.3 &                    4.4 &                         6.7 &                        8.6 &                        25.9 \\
hy       &          2.0 &        266.5 &          1.3 &        141.2 &          47.0 &                   13.0 &                  13.6 &                   18.1 &                         8.1 &                       13.4 &                        35.6 &                    5.3 &                   6.2 &                    5.5 &                         7.0 &                        9.0 &                        26.3 \\
id       &          1.1 &        140.8 &          1.1 &        120.7 &          14.3 &                    4.0 &                   4.3 &                    9.4 &                         5.9 &                        9.2 &                        27.6 &                    3.5 &                   4.1 &                    3.4 &                         6.6 &                        8.3 &                        25.3 \\
ig       &          1.2 &        159.1 &          1.3 &        137.4 &          13.7 &                    5.8 &                   5.7 &                   11.1 &                         6.0 &                       10.5 &                        28.1 &                    4.8 &                   5.8 &                    4.8 &                         6.7 &                        9.0 &                        25.9 \\
is       &          1.1 &        141.8 &          1.1 &        124.1 &          12.4 &                    4.9 &                   5.1 &                    9.7 &                         5.7 &                        9.1 &                        27.6 &                    4.2 &                   4.9 &                    4.1 &                         6.4 &                        8.6 &                        26.0 \\
it       &          1.2 &        155.4 &          1.2 &        130.9 &          15.8 &                    4.5 &                   4.9 &                   10.0 &                         6.1 &                        9.3 &                        28.1 &                    3.9 &                   4.7 &                    3.4 &                         6.4 &                        8.5 &                        25.9 \\
ja       &          1.3 &        165.1 &          1.4 &        151.6 &           8.2 &                    5.4 &                   5.0 &                    7.0 &                         6.3 &                       10.0 &                        28.4 &                    4.9 &                   7.5 &                    5.1 &                         6.7 &                        9.2 &                        26.5 \\
jv       &          1.0 &        135.6 &          1.1 &        117.2 &          13.5 &                    3.9 &                   4.3 &                    9.2 &                         5.6 &                        8.9 &                        27.6 &                    3.5 &                   4.0 &                    3.4 &                         6.4 &                        7.9 &                        25.3 \\
ka       &          2.9 &        385.0 &          1.4 &        154.4 &          59.9 &                   23.5 &                  16.0 &                   24.5 &                        10.3 &                       18.9 &                        44.9 &                    6.0 &                   7.9 &                    6.6 &                         7.1 &                        9.5 &                        28.3 \\
kk       &          1.9 &        247.0 &          1.2 &        132.1 &          46.5 &                   12.2 &                   9.7 &                   15.0 &                         7.7 &                       12.6 &                        33.8 &                    4.4 &                   4.8 &                    4.3 &                         6.3 &                        8.7 &                        27.6 \\
km       &          3.3 &        430.0 &          1.5 &        167.3 &          61.1 &                   27.6 &                  22.7 &                   29.0 &                        10.8 &                       20.6 &                        48.3 &                    7.0 &                  12.3 &                    9.3 &                         7.2 &                       10.1 &                        28.3 \\
kn       &          2.8 &        371.0 &          1.3 &        139.7 &          62.4 &                   34.1 &                  20.5 &                   23.5 &                         9.8 &                       18.4 &                        43.7 &                    4.8 &                   6.7 &                    5.7 &                         6.7 &                        9.1 &                        26.4 \\
ko       &          1.2 &        155.9 &          1.2 &        133.0 &          14.7 &                    4.5 &                   5.0 &                    9.1 &                         6.0 &                        9.7 &                        30.6 &                    4.7 &                   5.8 &                    4.0 &                         6.8 &                        8.7 &                        26.2 \\
ku       &          1.1 &        143.0 &          1.2 &        131.6 &           8.0 &                    4.7 &                   5.0 &                   10.1 &                         5.7 &                        9.1 &                        27.6 &                    4.5 &                   5.2 &                    4.4 &                         6.5 &                        8.8 &                        26.2 \\
ky       &          1.9 &        247.3 &          1.2 &        129.5 &          47.6 &                   12.2 &                   9.8 &                   14.6 &                         7.6 &                       12.9 &                        34.1 &                    4.3 &                   4.7 &                    4.1 &                         6.5 &                        8.6 &                        26.2 \\
lb       &          1.1 &        150.1 &          1.2 &        130.2 &          13.2 &                    4.7 &                   5.2 &                   10.8 &                         5.8 &                        9.3 &                        28.0 &                    4.3 &                   5.1 &                    4.1 &                         6.5 &                        8.7 &                        26.0 \\
lo       &          2.7 &        356.5 &          1.2 &        125.9 &          64.7 &                   30.8 &                  19.6 &                   22.7 &                         9.3 &                       17.5 &                        44.9 &                    4.6 &                   6.1 &                    5.2 &                         6.5 &                        8.6 &                        26.4 \\
lt       &          1.1 &        137.6 &          1.2 &        125.8 &           8.5 &                    4.4 &                   4.7 &                    9.2 &                         5.9 &                        8.7 &                        27.3 &                    4.5 &                   5.1 &                    4.2 &                         6.4 &                        8.6 &                        27.6 \\
lv       &          1.1 &        144.9 &          1.2 &        126.1 &          13.0 &                    4.8 &                   5.0 &                    9.6 &                         5.8 &                        9.2 &                        27.9 &                    4.6 &                   5.0 &                    4.3 &                         6.7 &                        8.5 &                        26.9 \\
mg       &          1.3 &        163.7 &          1.3 &        142.4 &          13.0 &                    5.5 &                   5.7 &                   11.1 &                         6.2 &                        9.8 &                        28.8 &                    4.6 &                   6.0 &                    4.8 &                         6.9 &                        9.0 &                        26.6 \\
mi       &          1.2 &        152.0 &          1.3 &        140.0 &           7.9 &                    4.8 &                   5.3 &                   10.7 &                         5.8 &                        9.9 &                        28.2 &                    4.7 &                   5.7 &                    4.6 &                         6.6 &                        9.0 &                        26.8 \\
mk       &          1.9 &        248.2 &          1.3 &        137.7 &          44.5 &                   12.2 &                   9.8 &                   14.7 &                         7.6 &                       12.9 &                        37.7 &                    4.3 &                   4.9 &                    3.9 &                         6.8 &                        8.8 &                        27.3 \\
ml       &          3.1 &        406.9 &          1.4 &        148.4 &          63.5 &                   37.6 &                  21.0 &                   25.9 &                        17.9 &                       19.4 &                        46.7 &                    5.4 &                   7.8 &                    6.5 &                         7.0 &                        9.2 &                        26.6 \\
mn       &          1.9 &        249.0 &          1.3 &        139.7 &          43.9 &                   12.2 &                  10.3 &                   15.0 &                         7.5 &                       13.5 &                        33.5 &                    4.9 &                   5.3 &                    4.7 &                         6.6 &                        8.9 &                        26.4 \\
mr       &          2.7 &        351.5 &          1.3 &        140.2 &          60.1 &                   26.8 &                  17.1 &                   22.9 &                         9.5 &                       17.0 &                        42.1 &                    5.1 &                   6.5 &                    5.0 &                         6.9 &                        9.0 &                        26.6 \\
ms       &          1.1 &        144.9 &          1.1 &        124.5 &          14.0 &                    4.2 &                   4.4 &                    9.7 &                         5.9 &                        8.9 &                        27.9 &                    3.6 &                   4.3 &                    3.6 &                         6.6 &                        8.3 &                        26.0 \\
mt       &          1.2 &        152.0 &          1.2 &        127.2 &          16.3 &                    5.2 &                   5.5 &                   11.0 &                         5.7 &                        9.9 &                        31.6 &                    4.4 &                   5.1 &                    4.3 &                         6.5 &                        8.6 &                        33.1 \\
my       &          3.5 &        460.0 &          1.2 &        136.1 &          70.4 &                   31.9 &                  21.3 &                   29.7 &                        11.6 &                       21.7 &                        51.4 &                    5.1 &                   6.9 &                    6.3 &                         7.0 &                        8.9 &                        26.4 \\
ne       &          2.6 &        335.4 &          1.2 &        130.3 &          61.2 &                   24.2 &                  16.1 &                   22.0 &                         9.1 &                       16.0 &                        40.7 &                    4.6 &                   5.5 &                    4.3 &                         7.5 &                        8.7 &                        26.2 \\
nl       &          1.1 &        146.0 &          1.1 &        125.5 &          14.1 &                    4.1 &                   4.8 &                    9.9 &                         6.0 &                        9.2 &                        27.8 &                    3.8 &                   4.5 &                    3.5 &                         6.4 &                        8.5 &                        25.9 \\
no       &          1.0 &        133.4 &          1.1 &        115.7 &          13.3 &                    3.8 &                   4.4 &                    9.1 &                         5.7 &                        9.0 &                        27.4 &                    3.5 &                   4.0 &                    3.3 &                         6.6 &                        8.2 &                        25.9 \\
ny       &          1.1 &        145.8 &          1.1 &        121.6 &          16.6 &                    4.7 &                   4.8 &                   10.3 &                         5.4 &                        9.6 &                        27.1 &                    3.9 &                   4.5 &                    3.9 &                         6.4 &                        8.4 &                        25.7 \\
pa       &          2.6 &        340.1 &          1.6 &        179.3 &          47.3 &                   26.9 &                  17.9 &                   23.8 &                         9.2 &                       16.7 &                        40.8 &                    7.4 &                  12.6 &                    8.6 &                         7.3 &                       10.5 &                        28.5 \\
pl       &          1.1 &        146.5 &          1.2 &        129.0 &          11.9 &                    4.7 &                   5.1 &                   10.1 &                         6.0 &                        9.4 &                        32.3 &                    4.5 &                   5.3 &                    4.3 &                         6.5 &                        8.4 &                        25.8 \\
ps       &          1.6 &        212.3 &          1.3 &        145.4 &          31.5 &                    8.4 &                   7.7 &                   14.7 &                         6.9 &                       11.9 &                        31.1 &                    4.6 &                   6.3 &                    5.0 &                         6.8 &                        9.1 &                        27.1 \\
pt       &          1.1 &        145.8 &          1.1 &        124.0 &          14.9 &                    4.1 &                   4.4 &                    9.3 &                         5.9 &                        9.4 &                        28.1 &                    3.8 &                   4.3 &                    3.3 &                         6.4 &                        8.4 &                        25.7 \\
ro       &          1.2 &        155.1 &          1.2 &        135.3 &          12.8 &                    4.8 &                   5.1 &                   10.4 &                         6.0 &                        9.5 &                        28.5 &                    4.5 &                   5.4 &                    4.3 &                         6.9 &                        8.8 &                        26.3 \\
ru       &          2.0 &        257.3 &          1.3 &        142.5 &          44.6 &                   12.5 &                   9.9 &                   14.5 &                         7.9 &                       13.0 &                        34.5 &                    4.6 &                   5.6 &                    3.8 &                         6.6 &                        9.0 &                        26.6 \\
sd       &          1.6 &        209.4 &          1.3 &        145.2 &          30.7 &                    8.1 &                   7.5 &                   14.8 &                         6.7 &                       12.0 &                        30.9 &                    4.8 &                   6.5 &                    5.3 &                         6.8 &                        9.2 &                        27.1 \\
si       &          2.6 &        342.0 &          1.4 &        149.2 &          56.4 &                   28.0 &                  18.2 &                   24.0 &                         9.2 &                       18.1 &                        40.9 &                    5.7 &                   7.5 &                    6.4 &                         6.8 &                        9.3 &                        27.0 \\
sk       &          1.1 &        142.0 &          1.1 &        124.7 &          12.2 &                    4.5 &                   4.9 &                    9.6 &                         5.9 &                        8.8 &                        27.6 &                    4.5 &                   4.9 &                    4.1 &                         6.3 &                        8.4 &                        23.2 \\
sl       &          1.0 &        132.4 &          1.1 &        117.5 &          11.2 &                    3.9 &                   4.5 &                    9.0 &                         5.8 &                        8.8 &                        27.7 &                    3.9 &                   4.3 &                    3.5 &                         6.6 &                        8.3 &                        27.3 \\
sm       &          1.2 &        158.2 &          1.3 &        146.3 &           7.5 &                    4.9 &                   5.6 &                   11.4 &                         5.8 &                       10.1 &                        28.2 &                    4.5 &                   6.2 &                    4.8 &                         6.7 &                        9.2 &                        26.2 \\
sn       &          1.1 &        145.9 &          1.1 &        124.1 &          14.9 &                    4.7 &                   4.7 &                   10.2 &                         5.7 &                        9.9 &                        27.8 &                    4.1 &                   4.9 &                    4.1 &                         6.5 &                        8.5 &                        26.5 \\
so       &          1.2 &        150.1 &          1.2 &        133.5 &          11.1 &                    4.8 &                   5.1 &                   11.0 &                         5.7 &                        9.8 &                        27.8 &                    4.4 &                   5.4 &                    4.5 &                         6.5 &                       13.2 &                        26.7 \\
sq       &          1.2 &        156.6 &          1.2 &        134.6 &          14.1 &                    5.4 &                   5.5 &                   10.8 &                         6.0 &                        9.6 &                        28.5 &                    4.7 &                   5.5 &                    4.6 &                         6.8 &                        8.8 &                        26.3 \\
sr       &          1.8 &        235.2 &          1.2 &        136.0 &          42.2 &                   11.1 &                   9.2 &                   13.8 &                         7.3 &                       12.5 &                        32.9 &                    4.3 &                   4.9 &                    4.0 &                         6.8 &                        8.8 &                        26.3 \\
st       &          1.2 &        157.2 &          1.3 &        142.9 &           9.1 &                    5.2 &                   5.4 &                   11.5 &                         5.6 &                       10.0 &                        27.8 &                    4.5 &                   6.0 &                    4.8 &                         6.7 &                        9.1 &                        26.4 \\
su       &          1.0 &        136.6 &          1.1 &        118.0 &          13.6 &                    3.9 &                   4.3 &                    9.2 &                         5.6 &                        8.9 &                        27.4 &                    3.6 &                   4.2 &                    3.5 &                         6.3 &                        8.3 &                        25.3 \\
sv       &          1.0 &        135.8 &          1.1 &        114.9 &          15.4 &                    3.9 &                   4.4 &                    9.2 &                         6.1 &                        8.7 &                        27.2 &                    3.7 &                   4.1 &                    3.3 &                         6.3 &                        8.3 &                        26.6 \\
sw       &          1.0 &        136.7 &          1.1 &        121.8 &          10.8 &                    4.1 &                   4.5 &                    9.6 &                         5.7 &                        8.9 &                        27.5 &                    3.8 &                   4.4 &                    3.6 &                         6.7 &                        8.1 &                        25.9 \\
ta       &          3.2 &        416.6 &          1.4 &        153.2 &          63.2 &                   39.4 &                  22.0 &                   24.9 &                        10.7 &                       19.8 &                        47.6 &                    5.5 &                   8.2 &                    6.7 &                         7.1 &                        9.4 &                        28.2 \\
te       &          2.7 &        349.5 &          1.3 &        140.3 &          59.9 &                   30.0 &                  17.9 &                   22.2 &                         9.4 &                       16.9 &                        41.9 &                    4.8 &                   6.5 &                    5.6 &                         6.9 &                        9.0 &                        26.5 \\
tg       &          2.0 &        262.4 &          1.4 &        150.4 &          42.7 &                   14.2 &                  11.0 &                   16.4 &                         7.8 &                       13.4 &                        34.7 &                    5.3 &                   6.5 &                    5.3 &                         7.0 &                        9.2 &                        26.6 \\
th       &          2.8 &        360.8 &          1.2 &        134.4 &          62.7 &                   30.5 &                  19.2 &                   20.5 &                         9.8 &                       17.5 &                        42.6 &                    4.6 &                   6.8 &                    5.1 &                         6.8 &                        8.8 &                        27.7 \\
tr       &          1.1 &        146.2 &          1.1 &        124.2 &          15.0 &                    4.8 &                   4.8 &                    9.5 &                         5.7 &                        9.4 &                        28.1 &                    4.2 &                   4.8 &                    4.1 &                         6.7 &                        8.4 &                        26.0 \\
uk       &          1.9 &        243.0 &          1.3 &        140.9 &          42.0 &                   11.7 &                   9.6 &                   14.2 &                         7.6 &                       12.8 &                        33.5 &                    4.7 &                   5.4 &                    4.3 &                         6.5 &                        8.9 &                        26.5 \\
ur       &          1.8 &        229.0 &          1.4 &        150.8 &          34.1 &                   10.1 &                   8.7 &                   15.7 &                         7.4 &                       12.3 &                        32.7 &                    4.9 &                   7.1 &                    5.4 &                         7.0 &                        9.4 &                        28.2 \\
uz       &          1.1 &        147.7 &          1.1 &        122.7 &          16.9 &                    4.8 &                   5.1 &                   10.7 &                         6.1 &                        9.0 &                        27.9 &                    4.1 &                   4.8 &                    4.0 &                         6.8 &                        8.4 &                        23.2 \\
vi       &          1.4 &        181.4 &          1.6 &        179.6 &           1.0 &                    7.1 &                   6.1 &                   11.7 &                         6.5 &                       10.6 &                        29.5 &                    7.3 &                  12.2 &                    8.3 &                         6.9 &                       10.2 &                        28.3 \\
xh       &          1.1 &        137.6 &          1.1 &        115.0 &          16.4 &                    4.4 &                   4.5 &                    9.9 &                         5.5 &                        9.4 &                        30.2 &                    3.9 &                   4.4 &                    3.8 &                         6.3 &                        8.2 &                        25.6 \\
yi       &          1.9 &        253.8 &          1.3 &        145.6 &          42.6 &                   12.5 &                  13.2 &                   18.1 &                         7.4 &                       13.5 &                        34.1 &                    5.0 &                   6.6 &                    5.7 &                         6.7 &                        9.2 &                        27.2 \\
yo       &          1.3 &        166.9 &          1.3 &        144.8 &          13.2 &                    6.3 &                   6.3 &                   11.9 &                         6.2 &                       10.0 &                        29.0 &                    5.3 &                   6.8 &                    5.4 &                         6.9 &                        9.3 &                        26.9 \\
zh       &          0.9 &        119.4 &          1.1 &        117.5 &           1.6 &                    3.4 &                   4.0 &                    6.0 &                         5.6 &                        8.6 &                        27.2 &                    3.4 &                   4.9 &                    3.3 &                         6.3 &                        8.2 &                        25.4 \\
zu       &          1.1 &        146.8 &          1.1 &        121.3 &          17.4 &                    4.8 &                   4.8 &                   10.6 &                         5.7 &                        9.9 &                        27.7 &                    4.1 &                   4.7 &                    4.1 &                         6.4 &                        8.4 &                        26.5 \\
\bottomrule
\end{tabular}
\caption{Results for each of the analyzed languages. The left-hand columns contain the comparison of enoding lengths \utf{} and \ours{}. The right-hand columns present performance (BPEB) and inference time of corresponding language models \byt{} and \myt{}. All numbers are averages across the FLORES-200 test split.}
\label{tab:results_per_language}
\end{table*}
}

%% file: tables/xtreme_up_qa.tex
\begin{table*}[!tb]
\centering
\small
\label{qa_in_lang_results}
\begin{tabular}{lccccccccccc}
\toprule
 &   ar &   bn &   en &   fi &   id &   ko &   ru &   sw &   te & AVG LR &  AVG \\
\midrule
ByT5 &  81.6 &  59.3 &  76.3 &  80.7 &  77.8 &  75.7 &  75.9 &  77.1 &  78.1 & 73.2 & 75.9 \\
MyT5 &  82.3 &  67.2 &  74.9 &  80.5 &  76.1 &  74.8 &  76.6 &  74.2 &  83.6 & 75.3 & 76.7 \\
\bottomrule
\end{tabular}
\caption{F1 scores for question answering (\xu{} benchmark).}
\label{tab:xtreme_up_qa}
\end{table*}

%% file: tables/xtreme_up_ner.tex
\begin{table*}[!tb]
\centering
\tiny
\resizebox{\textwidth}{!}{
\begin{tabular}{lcccccccccccccccccccccc}
\toprule
 &  am &  bbj &  bm &  ee &  ha &  ig &  lg &  luo &  mos &  ny &  pcm &  rw &  sn &  sw &  tn &  tw &  wo &  xh &  yo &  zu & AVG LR &  AVG \\
\midrule
ByT5 & 60.8 & 72.5 & 80.0 & 88.1 & 88.1 & 84.3 & 84.6 & 77.1 & 73.5 & 89.1 & 85.2 & 76.7 & 90.0 & 88.5 & 85.6 & 77.6 & 80.2 & 83.4 & 78.7 & 85.2 & 81.5 & 81.5 \\
MyT5 & 62.1 & 68.9 & 79.2 & 87.1 & 87.5 & 83.3 & 83.6 & 75.4 & 75.4 & 88.0 & 85.2 & 77.9 & 90.2 & 88.9 & 84.7 & 77.7 & 75.0 & 82.0 & 79.4 & 83.8 & 80.8 & 80.8 \\
\bottomrule
\end{tabular}
}
\caption{F1 scores for named entity recognition (MasakhaNER test set \citet{adelani-etal-2022-masakhaner} via \xu{} benchmark)}
\label{tab:xtreme_up_ner}
\end{table*}

%% file: tables/xtreme_up_sp.tex
\begin{table*}[!tb]
\centering
\tiny
\resizebox{\textwidth}{!}{
\begin{tabular}{lccccccccccccccccccccc}
\toprule
 &   am &   be &   bn &   de &   en &   es &   fi &   fr &   ha &   hi &   ja &  pt\_br &   ru &   sw &   ta &   th &   tr &   yo &   zu & AVG LR &  AVG \\
\midrule
ByT5 & 18.6 & 31.7 & 30.7 & 34.5 & 35.1 & 33.1 & 30.0 & 34.8 & 25.7 & 25.7 & 31.4 &   34.7 & 35.7 & 26.4 & 26.4 & 24.6 & 32.1 & 18.6 & 22.8 & 25.1 & 29.1 \\
MyT5 & 16.5 & 26.2 & 20.6 & 31.6 & 31.6 & 28.1 & 25.7 & 28.1 & 21.7 & 18.7 & 18.1 &   30.4 & 32.8 & 21.4 & 21.2 & 19.2 & 25.7 & 13.0 & 16.7 & 19.7 & 23.5 \\
\bottomrule
\end{tabular}
}
\caption{Exact match score for semantic parsing (\xu{} benchmark)}
\label{tab:xtreme_up_sp}
\end{table*}

%% file: tables/xtreme_up_mt.tex
\begin{table*}[!tb]
\centering
\small
\resizebox{\textwidth}{!}{
\begin{tabular}{lccccccccccccccccc}
\toprule
 &  am &  de &  el &  fr &  hy &  ja &  kk &  ko &  mt &  pl &  ru &  sn &  ta &  te & vi &  AVG LR &  AVG \\
\midrule
MyT5 &    9.4 &   31.9 &   21.9 &   36.5 &   22.6 &    9.2 &   20.1 &    7.7 &   26.2 &   25.3 &   24.3 &   27.8 &   22.5 &   18.4 &   24.1 &    21.1 & 21.8 \\
ByT5 &    8.8 &   35.4 &   22.1 &   41.8 &   22.8 &    9.3 &   20.9 &    7.1 &   34.1 &   27.8 &   26.9 &   27.4 &   21.3 &   17.4 &   23.9 &    21.9 & 23.1 \\
\bottomrule
\end{tabular}
}
\caption{ChrF scores for machine translation (Florers 200 test set \citet{nllb2022} via \xu{} benchmark).}
\label{tab:xtreme_up_mt}
\end{table*}

%% file: sections/93_technical.tex
\section{Technical Details}
\label{sec:appendix-technical}

\subsection{Compuatational Infrastracture}

The \myt{} and reimplemented \byt{} models were trained on TPUs available through Google Cloud Platform.
We used v3-8 for training small and base models and v3-32 for the large model.
The training took approximately 90h for small, 230h for base, and 190h for large models.
We are thankful to Google for providing free quotas for those machines through the TPU Research Cloud program.

The inference in language modeling experiments was run on an A40 GPU core.

\subsection{Fine-Tuning}

For few-shot fine-tuning, we choose the same hyperparameters and optimization strategy as in \citet{ruder_xtreme-up_2023}: $0.1$ dropout, $1e^{-3}$ learning rate with inverse square root decay.
The batch size was chosen to facilitate training on v3-8 TPU, specifically 128 for NER; 64 for MT, QA,  and semantic parsing.
The number of fine-tuning steps corresponded to the sizes of the training datasets: QA 6500, NER 6000, semantic parsing 1000, and MT 10000.
For machine translation, we selected the following sample of language both for training and evaluation: Telugu, Tamil, Greek, Armenian, Russian, Kazakh, Amharic, Vietnamese, Japanese, French, Korean, German, Marathi, and Polish.